\documentclass{article}

\usepackage{microtype}
\usepackage{graphicx}
\usepackage{subfigure}
\usepackage{booktabs} % for professional tables

\usepackage{epsfig}
\usepackage{graphicx}
\usepackage{amsmath}
\usepackage{amssymb}
\usepackage{multirow}
\usepackage{enumerate}
\usepackage{bm}
\usepackage{wrapfig}
\usepackage{epstopdf}

\usepackage{hyperref}

\usepackage[accepted]{icml2021}

\icmltitlerunning{Neural-Pull: Learning Signed Distance Functions from Point Clouds by Learning to Pull Space onto Surfaces}

\begin{document}

\twocolumn[
\icmltitle{Neural-Pull: Learning Signed Distance Functions from Point Clouds by Learning to Pull Space onto Surfaces}



\icmlsetsymbol{equal}{*}

\begin{icmlauthorlist}
\icmlauthor{Baorui Ma}{equal,to}
\icmlauthor{Zhizhong Han}{equal,goo}
\icmlauthor{Yu-Shen Liu}{to}
\icmlauthor{Matthias Zwicker}{ed}
%\icmlauthor{Aeiau Zzzz}{equal,to}
%\icmlauthor{Bauiu C.~Yyyy}{equal,to,goo}
%\icmlauthor{Cieua Vvvvv}{goo}
%\icmlauthor{Iaesut Saoeu}{ed}
%\icmlauthor{Fiuea Rrrr}{to}
%\icmlauthor{Tateu H.~Yasehe}{ed,to,goo}
%\icmlauthor{Aaoeu Iasoh}{goo}
%\icmlauthor{Buiui Eueu}{ed}
%\icmlauthor{Aeuia Zzzz}{ed}
%\icmlauthor{Bieea C.~Yyyy}{to,goo}
%\icmlauthor{Teoau Xxxx}{ed}
%\icmlauthor{Eee Pppp}{ed}
\end{icmlauthorlist}

\icmlaffiliation{to}{School of Software, BNRist, Tsinghua University, Beijing 100084, P. R. China}
\icmlaffiliation{goo}{Department of Computer Science, Wayne State University, Detroit, USA}
\icmlaffiliation{ed}{Department of Computer Science, University of Maryland, College Park, USA}

\icmlcorrespondingauthor{Yu-Shen Liu}{liuyushen@tsinghua.edu.cn}
%\icmlcorrespondingauthor{Eee Pppp}{ep@eden.co.uk}

\icmlkeywords{Machine Learning, ICML}

\vskip 0.3in
]



\printAffiliationsAndNotice{\icmlEqualContribution} % otherwise use the standard text.

\begin{abstract}
Reconstructing continuous surfaces from 3D point clouds is a fundamental operation in 3D geometry processing. Several recent state-of-the-art methods address this problem using neural networks to learn signed distance functions (SDFs). In this paper, we introduce \textit{Neural-Pull}, a new approach that is simple and leads to high quality SDFs. Specifically, we train a neural network to pull query 3D locations to their closest points on the surface using the predicted signed distance values and the gradient at the query locations, both of which are computed by the network itself. The pulling operation moves each query location with a stride given by the distance predicted by the network. Based on the sign of the distance, this may move the query location along or against the direction of the gradient of the SDF. This is a differentiable operation that allows us to update the signed distance value and the gradient simultaneously during training. Our outperforming results under widely used benchmarks demonstrate that we can learn SDFs more accurately and flexibly for surface reconstruction and single image reconstruction than the state-of-the-art methods. Our code and data are available at \url{https://github.com/mabaorui/NeuralPull}.
\end{abstract}
\section{Introduction}
Signed Distance Functions (SDFs) have been an important 3D shape representation for deep learning based 3D shape analysis~\cite{Park_2019_CVPR,MeschederNetworks,mildenhall2020nerf,DBLP:journals/corr/abs-1901-06802,pifuSHNMKL19,rematasICML21,sitzmann2019siren,DBLP:journals/corr/abs-2011-10379,takikawa2021nglod,DBLP:journals/corr/abs-2105-02788,DBLP:journals/corr/abs-2104-10078,DBLP:journals/corr/abs-2104-04532,DBLP:journals/corr/abs-2102-04776}, due to their advantages over other representations in representing high resolution shapes with arbitrary topology. Given ground truth signed distance values, it is intuitive to learn an SDF by training a deep neural network to regress signed distance values for query 3D locations, where an image~\cite{DBLP:journals/corr/abs-1901-06802,Park_2019_CVPR} or a point cloud~\cite{jia2020learning,ErlerEtAl:Points2Surf:ECCV:2020} representing the shape can serve as a condition which is an additional input of the network. It has also been shown how to learn SDFs from multiple 2D images rather than 3D information using differentiable renderers~\cite{DIST2019SDFRcvpr,Jiang2019SDFDiffDRcvpr,prior2019SDFRcvpr,DBLP:journals/cgf/WuS20}. In this paper, we address the problem of learning SDFs from raw point clouds and propose a new method that outperforms the state-of-the-art on widely used benchmarks.

Current solutions~\cite{gropp2020implicit,chibane2020neural,Atzmon_2020_CVPR,atzmon2020sald} aim to estimate unsigned distance fields by leveraging additional constraints. The rationale behind these solutions is that an unsigned distance field can be directly learned from the distances between a set of query 3D locations and their nearest neighbors on the 3D point clouds, while the signs of these distances require more information to infer, such as geometric regularization~\cite{gropp2020implicit}, sign agnostic learning~\cite{Atzmon_2020_CVPR,atzmon2020sald}, or analytical gradients~\cite{chibane2020neural}.

%This fact makes these solutions heavily rely on the point clouds, even during testing stage, which significantly limits their 3D representation ability in applications without knowing point clouds during inference, such as single image reconstruction.

In this paper, we propose a method to learn SDFs directly from raw point clouds without requiring ground truth signed distance values. Our method learns the SDF from a point cloud, or from multiple point clouds with conditions by training a neural network to learn to pull the surrounding 3D space onto the surface represented by the point cloud. Hence we call our method \emph{Neural-Pull}. Specifically, given a 3D query location as input to the network, we ask the network to pull it to its closest point on the surface using the predicted signed distance value and the gradient at the query location, both of which are calculated by the network itself. The pulling operation is differentiable, and depending on the sign of the predicted distance, it moves the query location along or against the direction of the gradient with a stride given by the signed distance. Since our training objective involves both the signed distance and its gradient, it leads to highly effective learning. Our experiments using widely used benchmarks show that Neural-Pull can learn SDFs more accurately and flexibly when representing 3D shapes in different applications than previous state-of-the-art methods. Our contributions are listed below.

\begin{enumerate}[i)]
\item We introduce Neural-Pull, a novel approach to learn SDFs directly from raw 3D point clouds without ground truth signed distance values.
\item We introduce the idea to effectively learn SDFs by updating the predicted signed distance values and the gradient simultaneously in order to pull surrounding 3D space onto the surface.
\item We significantly improve the state-of-the-art accuracy in surface reconstruction and single image reconstruction under various benchmarks.
\end{enumerate}

\section{Related Work}
Deep learning models have been playing an important role in different 3D computer vision applications~\cite{Zhizhong2018seq,wenxin_2020_CVPR,seqxy2seqzeccv2020paper,MAPVAE19,p2seq18,hutaoaaai2020,wenxin_2020_CVPR,Groueix_2018_CVPR,Tretschk2020PatchNets,bednarik2020,3DViewGraph19,tancik2020fourfeat,Zhizhong2018VIP,Han2019ShapeCaptionerGCacmmm,Badki_2020_CVPR,Mi_2020_CVPR,handrwr2020,wenxin_2021a_CVPR,wenxin_2021b_CVPR,Jiang2019SDFDiffDRcvpr,9187572,9318534}. In the following, we will briefly review work related to learning implicit functions for 3D shapes in different ways.

\noindent\textbf{Learning from 3D Ground Truth Globally. }Some techniques aim to learn implicit functions that represent conditional mappings from a 3D location to a binary occupancy value~\cite{MeschederNetworks,chen2018implicit_decoder} or a signed distance value~\cite{DBLP:journals/corr/abs-1901-06802,Park_2019_CVPR}. Early work requires the ground truth occupancy values or signed distance values as 3D supervision. For single image reconstruction, a single image~\cite{xu2019disn,pifuSHNMKL19,DBLP:conf/cvpr/ChibaneAP20,Gidi_2019_ICCV,Genova:2019:LST,seqxy2seqzeccv2020paper} or a learnable latent code~\cite{Park_2019_CVPR} can be a condition to provide information about a specified shape. For surface reconstruction~\cite{Williams_2019_CVPR,liu2020meshing,Mi_2020_CVPR,Genova:2019:LST}, we can leverage a point cloud as a condition to learn an implicit function which further produces a surface~\cite{jia2020learning,ErlerEtAl:Points2Surf:ECCV:2020}.

\noindent\textbf{Learning from 3D Ground Truth Locally. }To improve the performance of learning implicit functions, a local strategy was also explored that focuses on more local shape information. Jiang et al.~\cite{jiang2020lig} introduced the local implicit grid to improve the scalability and generality. Similarly, PatchNet~\cite{Tretschk2020PatchNets} was proposed to learn a patch-based surface representation to get more generalizable models. With a grid of independent latent codes, deep local shapes~\cite{DBLP:conf/eccv/ChabraLISSLN20} was proposed to represent 3D shapes without prohibitive memory requirements. Using locally interpolated features, convolutional occupancy networks~\cite{Peng2020ECCV} learn occupancy network for 3D scene reconstruction. Other local deep implicit functions~\cite{Genova_2020_CVPR} are learned by inferring the space decomposition and local deep implicit function learning from a 3D mesh or posed depth images.

\noindent\textbf{Learning from 2D Supervision. }We can also learn implicit functions from 2D supervision, such as multiple images. Vincent et al.~\cite{sitzmann2019srns} learned a mapping from world coordinates to a feature representation of local scene properties, which reduces the computational cost on sampling points for implicit surface learning. Inspired by ray marching rendering, different differentiable renderers~\cite{DIST2019SDFRcvpr,Jiang2019SDFDiffDRcvpr,prior2019SDFRcvpr} were introduced to render signed distance functions into images. In addition, ray-based field probing~\cite{shichenNIPS} or aggregating detection points on rays~\cite{DBLP:journals/cgf/WuS20} were employed to mine supervision for 3D occupancy fields. With the implicit differentiation, Niemeyer et al.~\cite{Volumetric2019SDFRcvpr} analytically derived in a differentiable rendering formulation for implicit shape and texture representations. For view synthesis, radiance fields were learned first, and then rendered using the differentiable volume rendering to calculate the loss~\cite{mildenhall2020nerf}.

\noindent\textbf{Learning from Point Clouds. }Without ground truth signed distance values or occupancy values, learning implicit functions directly from raw point clouds is more challenging. Current methods learn signed or unsigned distance fields with additional constraints, such as geometric regularization~\cite{gropp2020implicit}, sign agnostic learning with a specially designed loss function~\cite{Atzmon_2020_CVPR} or constraints on gradients~\cite{atzmon2020sald}, and analytical gradients~\cite{chibane2020neural}. A recent cocurrent work~\cite{chibane2020neural} learns to predict unsigned distances and infers the surface by pulling sampled queries to the surface. While our method directly learns SDFs which can be used to directly predict 3D shapes during testing, especially for applications without knowing point clouds during inference, such as single image reconstruction.

%sharing a similar idea with ours learns unsigned distance fields, which still requires direction to move a query location during the testing stage. While our method directly learns SDFs which can be used to directly predict 3D shapes during testing, especially for applications without knowing point clouds during inference, such as single image reconstruction.

\section{Method}
\noindent\textbf{Problem Statement. }We employ a neural network to learn SDFs that represent 3D shapes. An SDF $\bm{f}$ predicts a signed distance value $s\in\mathbb{R}$ for a query 3D location $\bm{q}=[x,y,z]$. Optionally, we provide an additional condition $\bm{c}$ as input, such that $\bm{f}(\bm{c},\bm{q})=s$. Given ground truth signed distances as supervision, current methods~\cite{DBLP:journals/corr/abs-1901-06802,Park_2019_CVPR} can employ a neural network to learn $\bm{f}$ as a regression problem. Different from these method, we aim to learn SDF $\bm{f}$ in a 3D space directly from 3D point cloud $\bm{P}=\{\bm{p}_j, j\in[1,J]\}$.

\noindent\textbf{Overview. }We introduce Neural-Pull as a neural network to learn how to pull a 3D space onto the surface represented by the point cloud $\bm{P}$. Rather than leveraging unsigned distances as previous methods~\cite{gropp2020implicit,chibane2020neural,Atzmon_2020_CVPR,atzmon2020sald}, Neural-Pull trains an SDF $\bm{f}$ to predict signed distances using the point cloud $\bm{P}$ and the gradient within the network itself to represent 3D shapes. Neural-pull tries to learn to pull a query location $\bm{q}_i$ which is randomly sampled around the surface to its nearest neighbor $\bm{t}_i$ on the surface, where the query locations form a set $\bm{Q}=\{\bm{q}_i, i\in[1,I]\}$. The pulling operation pulls the query location $\bm{q}_i$ with a stride of signed distance $s_i$, along or against the direction of the gradient $\bm{g}_i$ at $\bm{q}_i$, obtained within the network.

\begin{figure}[tb]
  \centering
  % the following command controls the width of the embedded PS file
  % (relative to the width of the current column)
  %\includegraphics[width=.95\linewidth, bb=39 696 126 756]{figures/definition3.eps}
   \includegraphics[width=0.75\linewidth]{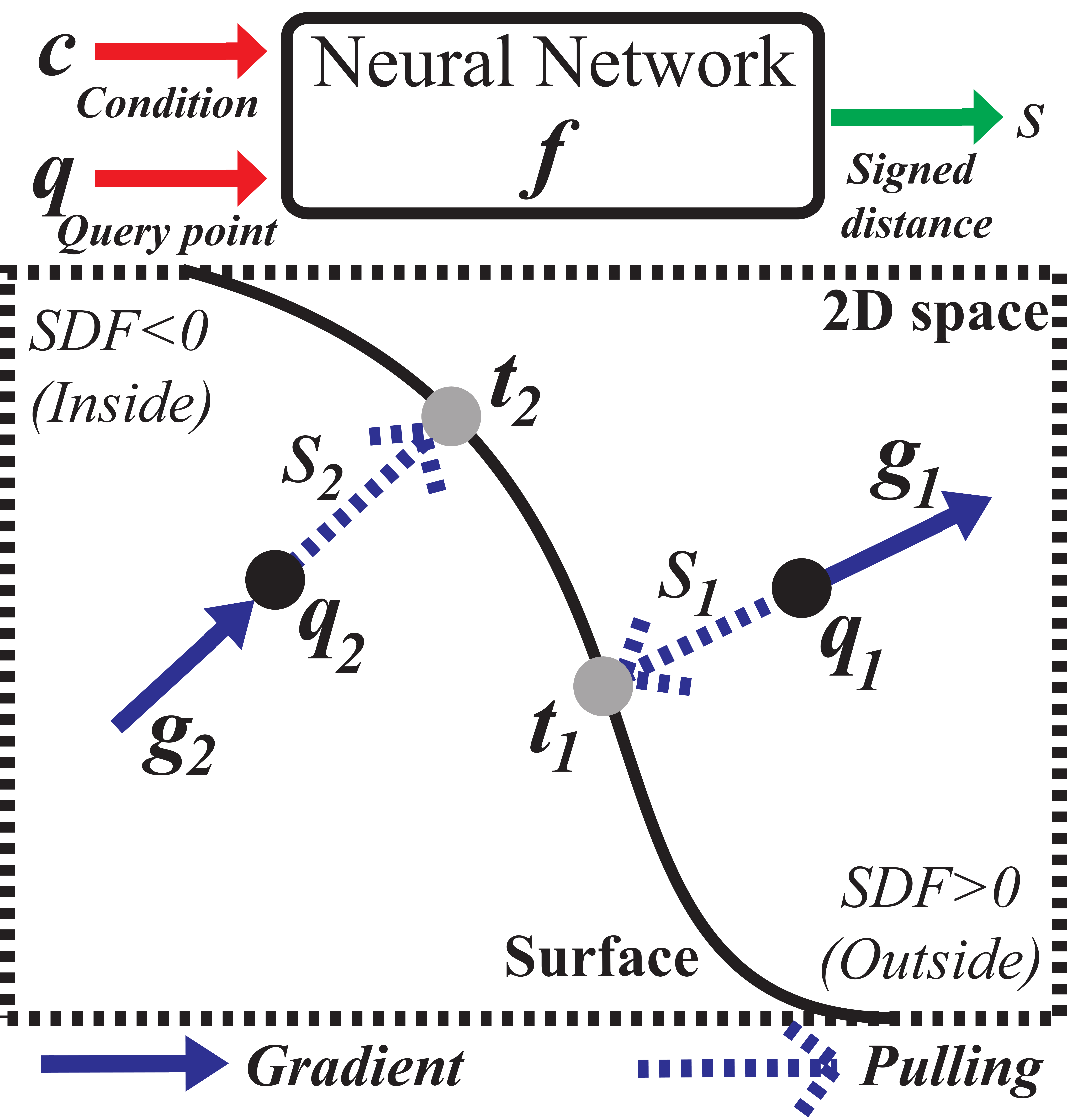}
  % replacing the above command with the one below will explicitly set
  % the bounding box of the PS figure to the rectangle (xl,yl),(xh,yh).
  % It will also prevent LaTeX from reading the PS file to determine
  % the bounding box (i.e., it will speed up the compilation process)
  % \includegraphics[width=.95\linewidth, bb=39 696 126 756]{sampleFig}
  %
  %
\caption{\label{fig:frameworks} Demonstration of pulling surrounding 2D space to a surface, where gradients $\bm{g}_i$ and signed distance value $s$ are from neural network $\bm{f}$. }
\end{figure}

We demonstrate our idea using a 2D surface in Fig.~\ref{fig:frameworks}, where the 2D surface splits the space into inside and outside of the shape. We train a neural network to employ the predicted signed distances $s_1$ (or $s_2$) to pull the query location $\bm{q}_1$ (or $\bm{q}_2$) to its nearest neighbor $\bm{t}_1$ (or $\bm{t}_2$ ) against (or along) the gradient $\bm{g}_1$ (or $\bm{g}_2$ ) at the query location $\bm{q}_1$ (or $\bm{q}_2$).

\noindent\textbf{Pulling Query Points. }We pull a 3D query location $\bm{q}_i$ to its nearest neighbor $\bm{t}_i$ on the surface using the predicted signed distance $s_i$ and the gradient $\bm{g}_i$ at $\bm{q}_i$ within the network. The gradient $\bm{g}_i$ is a vector whose components are the partial derivatives of $\bm{f}$ at $\bm{q}_i$, such that $\bm{g}_i=[\partial \bm{f}(\bm{c},\bm{q}_i)/\partial x, \partial \bm{f}(\bm{c},\bm{q}_i)/\partial y, \partial \bm{f}(\bm{c},\bm{q}_i)/\partial z]$, which is also denoted as $\nabla\bm{f}(\bm{c},\bm{q}_i)$, where $\bm{c}$ is a condition. It is the direction of the fastest signed distance increase in 3D space. Therefore, we can leverage this property to move a query location along or against the direction of gradient $\bm{g}_i$ to its nearest point on the surface. We leverage the following equation to pull query locations $\bm{q}_i$,

\begin{equation}
\label{eq:cd}
\begin{aligned}
\bm{t}'_i=\bm{q}_i-\bm{f}(\bm{c},\bm{q}_i)\times\nabla\bm{f}(\bm{c},\bm{q}_i)/||\nabla\bm{f}(\bm{c},\bm{q}_i)||_2,
\end{aligned}
\end{equation}

\noindent where $\bm{t}'_i$ is the pulled query location $\bm{q}_i$ after pulling, $\bm{c}$ is the condition to represent ground truth point cloud $\bm{P}$, and $\nabla\bm{f}(\bm{c},\bm{q}_i)/||\nabla\bm{f}(\bm{c},\bm{q}_i)||_2$ is the direction of gradient $\nabla\bm{f}(\bm{c},\bm{q}_i)$. Since $\bm{f}$ is a continuously differentiable
function, we can obtain $\nabla\bm{f}(\bm{c},\bm{q}_i)$ in the back-propagation process of training $\bm{f}$. As Fig.~\ref{fig:frameworks} demonstrates, for query locations inside of the shape $\bm{P}$, if the sign of the signed distance value is negative, and the network will move the query location $\bm{q}_i$ along the direction of gradient to $\bm{t}'_i$ on $\bm{P}$ using $\bm{t}'_i=\bm{q}_i+|\bm{f}(\bm{c},\bm{q}_i)|\times\nabla\bm{f}(\bm{c},\bm{q}_i)/||\nabla\bm{f}(\bm{c},\bm{q}_i)||_2$. Instead, the network will move query locations outside of $\bm{P}$ against the direction of gradient due to the positive signed distance value, using $\bm{t}'_i=\bm{q}_i-|\bm{f}(\bm{c},\bm{q}_i)|\times\nabla\bm{f}(\bm{c},\bm{q}_i)/||\nabla\bm{f}(\bm{c},\bm{q}_i)||_2$.

\noindent\textbf{Query Locations Sampling. }We randomly sample query locations around each point $\bm{p}_j$ of the ground truth point cloud $\bm{P}$. For each point $\bm{p}_j\in\bm{P}$, we establish an isotropic Gaussian function $\mathcal{N}(\bm{p}_j,\sigma^2)$ to form a distribution, according to which we randomly sample 25 query locations, where $\sigma^2$ is the parameter to control how far away from the surface we can sample query locations. Here, we employ an adaptive way to set $\sigma^2$  as the square distance between $\bm{p}_j$ and its $50$-th nearest neighbor, which reflects location density around $\bm{p}_j$. The sampled query locations can cover the area around the surface represented by the point cloud $\bm{P}$, both inside and outside of the shape. Our preliminary results show that sampling near the surface will improve the learning accuracy, since it is hard for the network to predict accurate signed distance and gradient to move a query location that is far from surface to its nearest neighbor on the surface. We will elaborate on the details of leveraging these query locations sampled around $\bm{P}$ during training later.

\noindent\textbf{Loss Function. }Neural-pull aims to train a network to learn to pull a query location $\bm{q}_i$ to its nearest neighbor $\bm{t}_i$ on the point cloud $\bm{P}$. So, we leverage a square error to minimize the distance between the pulled query location $\bm{t}'_i$ obtained in Eq.~\ref{eq:cd} and the nearest neighbor $\bm{t}_i$ among $\bm{p}_j$ on $\bm{P}$ below,

\begin{equation}
\label{eq:cd1}
%\begin{aligned}
d(\{\bm{t}'_i\},\{\bm{t}_i\})=\frac{1}{I}\sum_{i\in[1,I]}||\bm{t}'_i-\bm{t}_i||_2^2,
%\end{aligned}
\end{equation}

%\noindent where $\bm{t}_i$ is the nearest neighbor of $\bm{q}_i$ on $\bm{P}$, such that $\bm{t}_i=\min_{\bm{p}\in\bm{P}}||\bm{q}_i-\bm{p}||_2^2$.

\begin{figure}[tb]
  \centering
  % the following command controls the width of the embedded PS file
  % (relative to the width of the current column)
  %\includegraphics[width=.95\linewidth, bb=39 696 126 756]{figures/definition3.eps}
   \includegraphics[width=\linewidth]{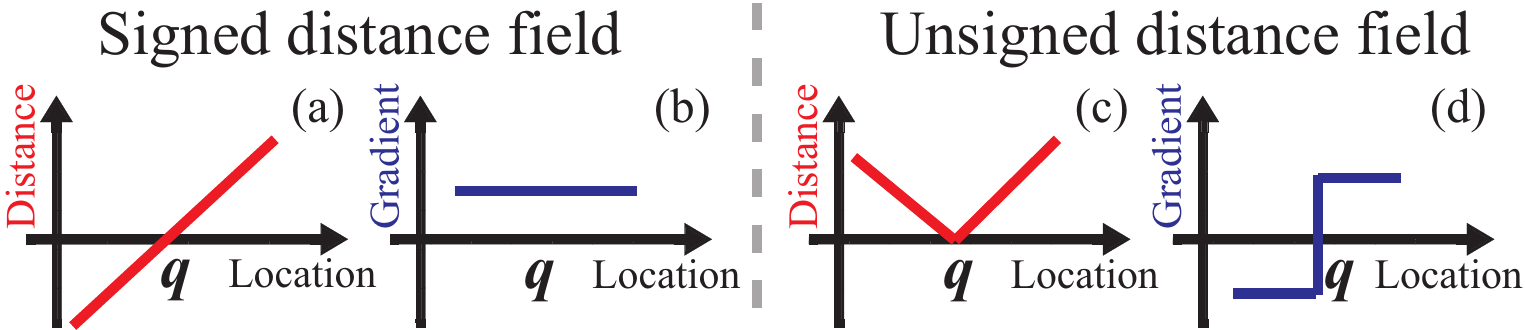}
  % replacing the above command with the one below will explicitly set
  % the bounding box of the PS figure to the rectangle (xl,yl),(xh,yh).
  % It will also prevent LaTeX from reading the PS file to determine
  % the bounding box (i.e., it will speed up the compilation process)
  % \includegraphics[width=.95\linewidth, bb=39 696 126 756]{sampleFig}
  %
  %
\caption{\label{fig:proof}  The illustration of the difference between signed distance field and unsigned distance field in terms of distance sign in (a), (c) and gradient in (b), (d).}
\end{figure}

\begin{figure}[tb]
  \centering
  % the following command controls the width of the embedded PS file
  % (relative to the width of the current column)
  %\includegraphics[width=.95\linewidth, bb=39 696 126 756]{figures/definition3.eps}
   \includegraphics[width=\linewidth]{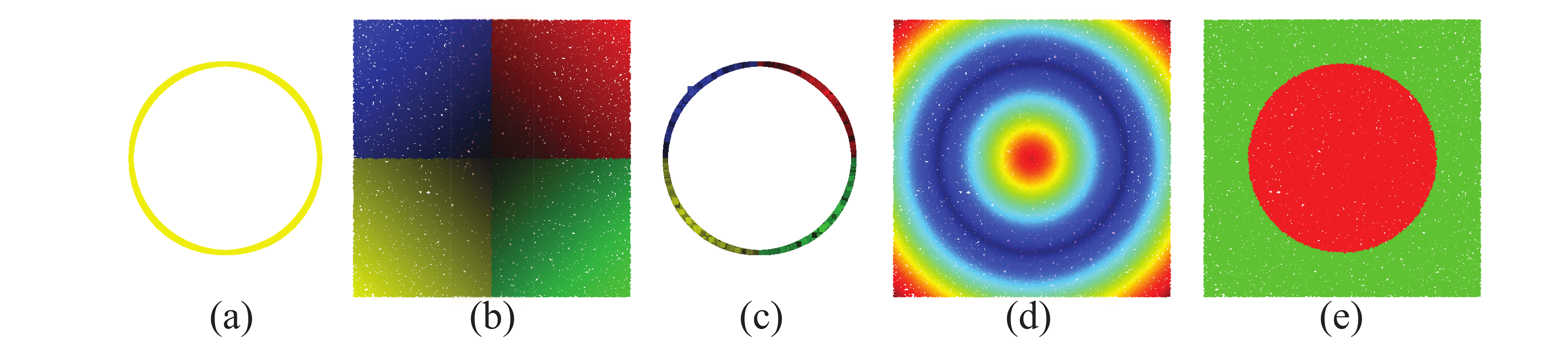}
  % replacing the above command with the one below will explicitly set
  % the bounding box of the PS figure to the rectangle (xl,yl),(xh,yh).
  % It will also prevent LaTeX from reading the PS file to determine
  % the bounding box (i.e., it will speed up the compilation process)
  % \includegraphics[width=.95\linewidth, bb=39 696 126 756]{sampleFig}
  %
  %
  %\vspace{-0.3in}
\caption{\label{fig:Teaser1}Optimization visualization on a 2D case.}
%\vspace{-0.18in}
\end{figure}

\noindent\textbf{Convergence to SDF. }One question that is not answered yet is that why the learned function $\bm{f}$ can converge to a signed distance field. Obviously, Eq.~\ref{eq:cd} is also satisfied in an unsigned distance field. We illustrate the difference between signed distance field and unsigned distance field near a 2D surface in Fig.~\ref{fig:proof}, where the location is shown in 1D. For a query point $\bm{q}$ on the surface, the signed distance field near $\bm{q}$, $\bm{q}+\bm{\triangle}\bm{q}$, is changing the sign of distance when going across the surface in Fig.~\ref{fig:proof}(a), while keeping the gradient the same in Fig.~\ref{fig:proof}(b). In contrast, the direction of the gradient for the region of $\bm{q}+\bm{\triangle}\bm{q}$ in an unsigned distance field is changing in Fig.~\ref{fig:proof}(d) since the unsigned distance is increasing in both sides of the surface in Fig.~\ref{fig:proof}(c). According to this difference, we have the following theorem indicating that a continuous function implemented by MLP can automatically converge to an SDF using our loss.

%According to this difference, we have the following theorem indicating that our network can converge to a signed distance function $\bm{f}$, since we introduce Eq.~\ref{eq:cd2} as a criterion.

%According to this difference, we have the following theorem indicating that our network can converge to a signed distance function $\bm{f}$.

%\noindent\textbf{Theorem 1. }\textit{Let $\bm{f}$ be a Multi-Layer Perceptron (MLP) with ReLU activation function in every layer except the last one. The MLP can converge to a signed distance function if Eq.~\ref{eq:cd2} is satisfied in any locations in the field.}
%
%\begin{equation}
%\label{eq:cd2}
%%\begin{aligned}
%\lim\limits_{\bm{\triangle}\bm{q}\to0}\nabla\bm{f}(\bm{c},\bm{q}+\bm{\triangle}\bm{q})=\nabla\bm{f}(\bm{c},\bm{q}).
%%\end{aligned}
%\end{equation}
%
%
%
%The proof to Theorem 1 is provided in the supplementary
%material.

\noindent\textbf{Theorem 1. }\textit{A continuous function $f$ implemented by MLP which is trained to minimize Eq.~\ref{eq:cd1} can converge to a signed distance function if Eq.~\ref{eq:cd2} is satisfied at any point $\bm{p}$ on the surface ($f(\bm{p})=0$), where $\bm{N}$ is the norm of $\bm{p}$, $\|\triangle t\|<\mu$ and $\mu$ indicates a small range.}

\begin{equation}
\label{eq:cd2}
%\begin{aligned}
\ f(\bm{p}-\bm{N}\triangle t)=-f(\bm{p}+\bm{N}\triangle t).
%\end{aligned}
\end{equation}

\noindent\textbf{Proof: }Since $f$ is a continuous function representing SDF, if $\nabla f(\bm{p})\neq \bm{0}$, we have $\bm{N}=\nabla f(\bm{p})/||\nabla f(\bm{p})||_2$. Assume $\triangle \bm{p}=\bm{N}\triangle t$, using the definition of gradient, we have

\begin{equation}
\label{eq:s1}
%\vspace{-0.1in}
\begin{aligned}
\lim_{\triangle \bm{p} \to \bm{0}}(f(\bm{p}+\triangle \bm{p})-f(\bm{p}))/\triangle \bm{p}=\bm{N}*||\nabla f(\bm{p})||_2.
\end{aligned}
\end{equation}

We can rewrite the equation above by removing $\lim$ into

\begin{equation}
\label{eq:s2}
%\vspace{-0.1in}
\begin{aligned}
(f(\bm{p}+\triangle \bm{p})-f(\bm{p}))/\triangle \bm{p}=\bm{N}*||\nabla f(\bm{p})||_2+\alpha,
\end{aligned}
\end{equation}

\noindent where $\alpha$ is infinitesimal when $\triangle \bm{p} \to \bm{0}$. We can further have $f(\bm{p}+\triangle \bm{p})-f(\bm{p})=(\bm{N}*||\nabla f(\bm{p})||_2+\alpha)*\triangle \bm{p}\neq 0$ by multiplying $\triangle \bm{p}$ on both sides, since $\triangle \bm{p}$ approaches $\bm{0}$ but never equals to $\bm{0}$. Similarly, we also have $f(\bm{p}-\triangle \bm{p})-f(\bm{p})=-(\bm{N}*||\nabla f(\bm{p})||_2+\alpha)*\triangle \bm{p}$. Since $f(\bm{p})=0$, we have

\begin{equation}
\label{eq:s3}
%\vspace{-0.1in}
\begin{aligned}
f(\bm{p}-\triangle \bm{p})=-f(\bm{p}+\triangle \bm{p}).
\end{aligned}
\end{equation}

\noindent We can further replace $\triangle \bm{p}$ into $\bm{N}\triangle t$ to get Eq.~\ref{eq:cd2} proofed.

Next, we will further proof our loss can significantly penalize $\nabla f(\bm{p})=\bm{0}$. Assume $\nabla f(\bm{p})=\bm{0}$, so $\lim_{\triangle \bm{p} \to \bm{0}}(f(\bm{p}+\triangle \bm{p})-f(\bm{p}))/\triangle \bm{p}=\bm{0}$. Since $f(\bm{p})=\bm{0}$, $f(\bm{p}+\triangle \bm{p})$ is higher order infinitesimal of $\triangle \bm{p}$. If we pull $\bm{p}+\triangle \bm{p}$ to $\bm{p}$, our loss is $||\bm{p}-(\bm{p}+\triangle \bm{p}-f(\bm{p}+\triangle \bm{p})\times \nabla f(\bm{p}+\triangle \bm{p})/||\nabla f(\bm{p}+\triangle \bm{p})||_2)||_2^2$, which can be rewritten into $||\triangle \bm{p}-f(\bm{p}+\triangle \bm{p})\times \nabla f(\bm{p}+\triangle \bm{p})/||\nabla f(\bm{p}+\triangle \bm{p})||_2||_2^2$. However, this equation can not be $\bm{0}$ since $f(\bm{p}+\triangle \bm{p})\times \nabla f(\bm{p}+\triangle \bm{p})/||\nabla f(\bm{p}+\triangle \bm{p})||_2$ is still higher order infinitesimal of $\triangle \bm{p}$. So, $\nabla f(\bm{p})\neq\bm{0}$.

\noindent\textbf{Optimization Visualization. }We demonstrate the optimization using a 2D case in Fig.~\ref{fig:Teaser1}. We learn a circle $\bm{P}$ in Fig.~\ref{fig:Teaser1} (a) using query locations $\bm{q}_i$ sampled in Fig.~\ref{fig:Teaser1} (b), where the color of $\bm{q}_i$ is used to track the pulled query locations $\bm{t}'_i$ in Fig.~\ref{fig:Teaser1} (c). The consistent color indicates that our loss can correctly pull the queries onto the surface. Additionally, we visualize the unsigned distances of the learned signed distance field in Fig.~\ref{fig:Teaser1} (d) and their signs in Fig.~\ref{fig:Teaser1} (e). Fig.~\ref{fig:Teaser1} justifies the effectiveness of our method.

\noindent\textbf{Training. }We randomly sample $J=2\times10^4$ points $\bm{p}_j$ from point clouds formed by $1\times10^5$ points released by OccNet~\cite{MeschederNetworks} as the ground truth point cloud $\bm{P}$ for each shape, where $j\in[1,J]$. As mentioned, we sample $25$ 3D query locations $\bm{q}_i$ around each point $\bm{p}_j$ to form the corresponding query location set $\bm{Q}$, such that $i\in[1,I]$ and $I=5\times10^5$. During training, we randomly select $5000$ query locations from $\bm{Q}$ as a batch to train the network. We try two different ways to select the $5000$ query locations. One way is to randomly select from $\bm{Q}$, the other is to uniformly sample $5000$ points on the ground truth point cloud $\bm{P}$, and then select one query location around each sampled point, where the second way can better cover the whole shape in each batch. Our preliminary results show that both of the two ways achieve good learning performance.

We employ a neural network similar to OccNet~\cite{MeschederNetworks} to learn the signed distance function (more details can be found in our supplemental material). We use the Adam optimizer with an initial learning rate of $0.0001$, and train the model in 2500 epochs. Moreover, we initialize the parameters in our network using the geometric network initialization (GNI)~\cite{Atzmon_2020_CVPR} to approximate the signed distance function of a sphere, where the sign of the signed distance inside of the shape is negative and positive outside.

\section{Experiments and Analysis}
%We evaluate our method by comparing it with state-of-the-art methods in surface reconstruction and single image reconstruction.

\subsection{Surface Reconstruction}
\noindent\textbf{Details. }We employ Neural-Pull to reconstruct 3D surfaces from point clouds. Given a point cloud $\bm{P}$, we do not leverage any condition $\bm{c}$ in Fig.~\ref{fig:frameworks} and overfit the neural network to the shape by minimizing the loss in Eq.~\ref{eq:cd1}, where we remove the network for extracting the feature of the condition. Hence our method does not require any training procedure under the training set, which differentiates our method from the previous ones~\cite{Atzmon_2020_CVPR,chibane2020neural,liu2020meshing,ErlerEtAl:Points2Surf:ECCV:2020}. After overfitting on each shape, our neural network learns an SDF for the shape. Then, we use the marching cubes~\cite{Lorensen87marchingcubes} algorithm to reconstruct the mesh surface.

\noindent\textbf{Dataset and Metric. }For fair comparison with other methods, we leverage three widely used benchmarks to evaluate our method in surface reconstruction.

\begin{table}[h]
\centering
\caption{Reconstruction comparison in terms of L2-CD ($\times100$).}  % ????????
\resizebox{\linewidth}{!}{
    \begin{tabular}{c|c|c|c|c|c|c}  % 32=0.1S£¬64=0.2S£¬128=1.3S * 600000
     \hline
          Dataset& DSDF & ATLAS & PSR& Points2Surf& IGR & Ours\\   % ?????§á?
     \hline
       ABC& 8.41 & 4.69 & 2.49& 1.80& 0.51 & \textbf{0.48} \\ %0.4806
       FAMOUS &10.08&4.69&1.67&1.41&1.65&\textbf{0.22} \\
     \hline
     Mean&9.25&4.69&2.08&1.61&1.08&\textbf{0.35}\\
     \hline
   \end{tabular}}
   \label{table:NOX1}
\end{table}

The first benchmark is the ABC dataset~\cite{Koch_2019_CVPR} which contains a large number and variety of CAD models. We use a subset of this dataset, released by Points2Surf~\cite{ErlerEtAl:Points2Surf:ECCV:2020} with the same train/test splitting.
%, which contains 4,950 clean watertight meshes for training and 100 meshes for testing.
The second benchmark is FAMOUS which is also released by Points2Surf~\cite{ErlerEtAl:Points2Surf:ECCV:2020}. The FAMOUS dataset is formed by 22 diverse well-known meshes.%that are well-known in geometry processing, such as the Utah teapot and the Stanford Bunny.
The last one is a subset of ShapeNet~\cite{shapenet2015}
%, which contains 23,108 CAD models in eight shape classes from ShapeNet.
%We employ this subset with
the same train/test splitting released by MeshingPoint (MeshP)~\cite{liu2020meshing}.

\begin{figure}[tb]
  \centering
  % the following command controls the width of the embedded PS file
  % (relative to the width of the current column)
  %\includegraphics[width=.95\linewidth, bb=39 696 126 756]{figures/definition3.eps}
   \includegraphics[width=\linewidth]{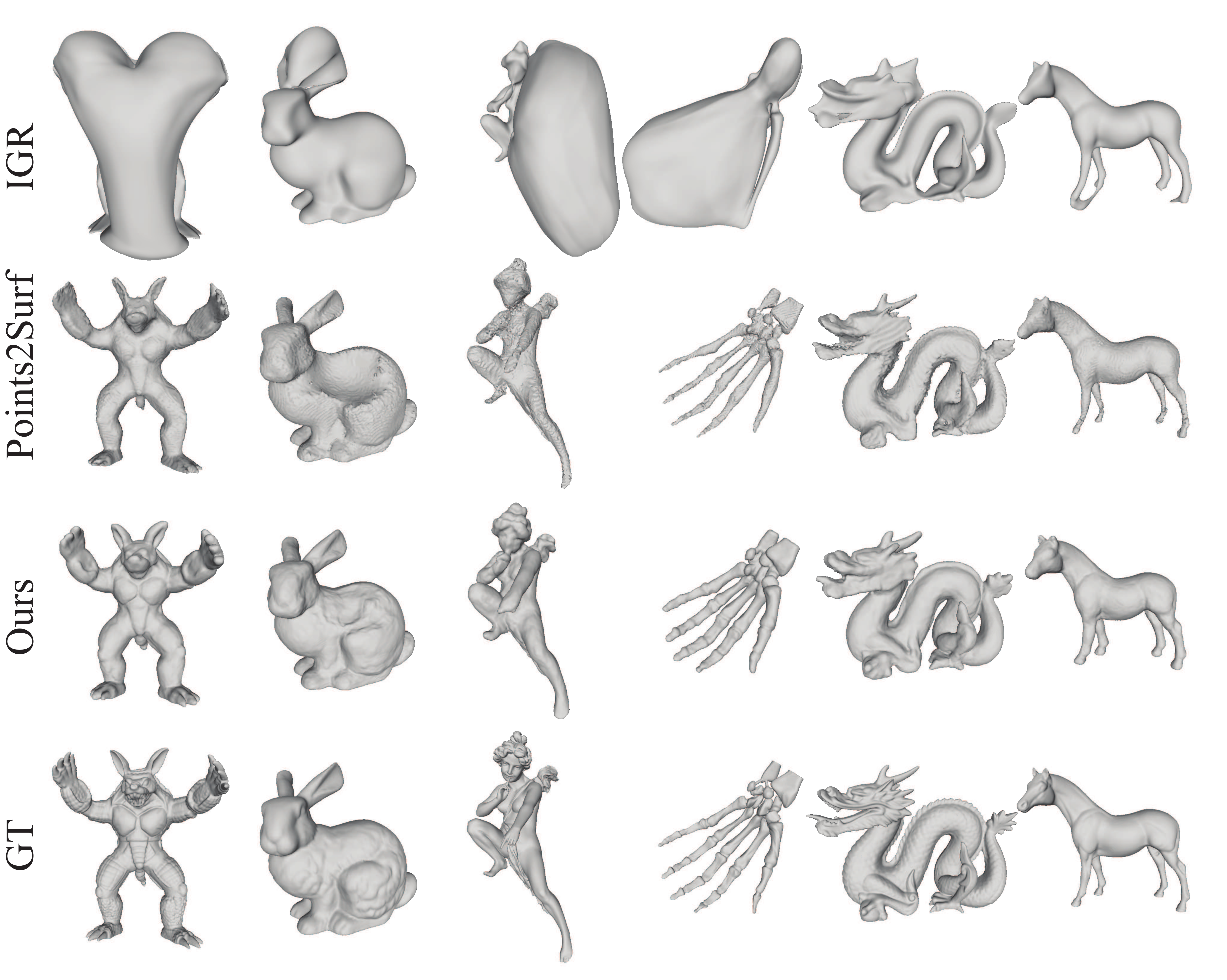}
  % replacing the above command with the one below will explicitly set
  % the bounding box of the PS figure to the rectangle (xl,yl),(xh,yh).
  % It will also prevent LaTeX from reading the PS file to determine
  % the bounding box (i.e., it will speed up the compilation process)
  % \includegraphics[width=.95\linewidth, bb=39 696 126 756]{sampleFig}
  %
  %
\caption{\label{fig:Famous} Comparison under FAMOUS in surface reconstruction.}
\end{figure}

\begin{table}[tb]
\centering
\caption{Surface reconstruction comparison in terms of L2-CD ($\times100$).}  % ????????Comparison of shape reconstruction with known camera pose from silhouette images with different resolutions in terms of CD
\resizebox{\linewidth}{!}{
    \begin{tabular}{|c|c|c|c|c|c|c|c|c|c|c|c|c|c|c|c}  % ?????
     \hline
       %\cline{1-12}
       %\hline
        Class& PSR& DMC & BPA & ATLAS &DMC&DSDF& DGP &MeshP&NUD&SALD&Ours \\  % ?????§á?
     \hline
        Display& 0.273& 0.269 & 0.093 & 1.094 &0.662&0.317& 0.293 & 0.069 & 0.077 &-& \textbf{0.039}\\
        Lamp &0.227&0.244&0.060&1.988&3.377&0.955&0.167&\textbf{0.053}&0.075&0.071&0.080\\
        Airplane&0.217&0.171&0.059&1.011&2.205&1.043&0.200&0.049&0.076&0.054&\textbf{0.008}\\
        Cabinet&0.363&0.373&0.292&1.661&0.766&0.921&0.237&0.112&0.041&-&\textbf{0.026}\\
        Vessel&0.254&0.228&0.078&0.997&2.487&1.254&0.199&0.061&0.079&-&\textbf{0.022}\\
        Table&0.383&0.375&0.120&1.311&1.128&0.660&0.333&0.076&0.067&0.066&\textbf{0.060}\\
        Chair&0.293&0.283&0.099&1.575&1.047&0.483&0.219&0.071&0.063&0.061&\textbf{0.054}\\
        Sofa&0.276&0.266&0.124&1.307&0.763&0.496&0.174&0.080&0.071&0.058&\textbf{0.012}\\
     \hline
     Mean&0.286&0.276&0.116&1.368&1.554&0.766&0.228&0.071&0.069&0.062&\textbf{0.038}\\
     \hline
   \end{tabular}}
   \label{table:t10}
\end{table}

To comprehensively evaluate our method with the state-of-the-art methods, we leverage different metrics for fair comparison. Following Points2Surf~\cite{ErlerEtAl:Points2Surf:ECCV:2020}, we leverage the L2-Chamfer distance (L2-CD) to evaluate the reconstruction error between our reconstruction and the $1\times10^4$ ground truth points under the ABC and FAMOUS datasets, where we also randomly sample $1\times10^4$ points on our reconstructed mesh. Besides the L1-Chamfer distance (L1-CD), we also follow MeshP~\cite{liu2020meshing} to leverage L2-CD, Normal Consistency (NC)~\cite{MeschederNetworks}, and F-score~\cite{Tatarchenko_2019_CVPR} to evaluate the reconstruction error, where we compare the $1\times10^5$ points sampled on the reconstructed shape with the $1\times10^5$ ground truth points released by OccNet~\cite{MeschederNetworks}. Note that L2-CD leverages the L2 norm to evaluate the distance between each pair of points, while L1-CD leverages the L1 norm. %We also indicate when we multiply by 100 to make the results readable in the caption of the tables.

\begin{figure}[tb]
  \centering
  % the following command controls the width of the embedded PS file
  % (relative to the width of the current column)
  %\includegraphics[width=.95\linewidth, bb=39 696 126 756]{figures/definition3.eps}
   \includegraphics[width=\linewidth]{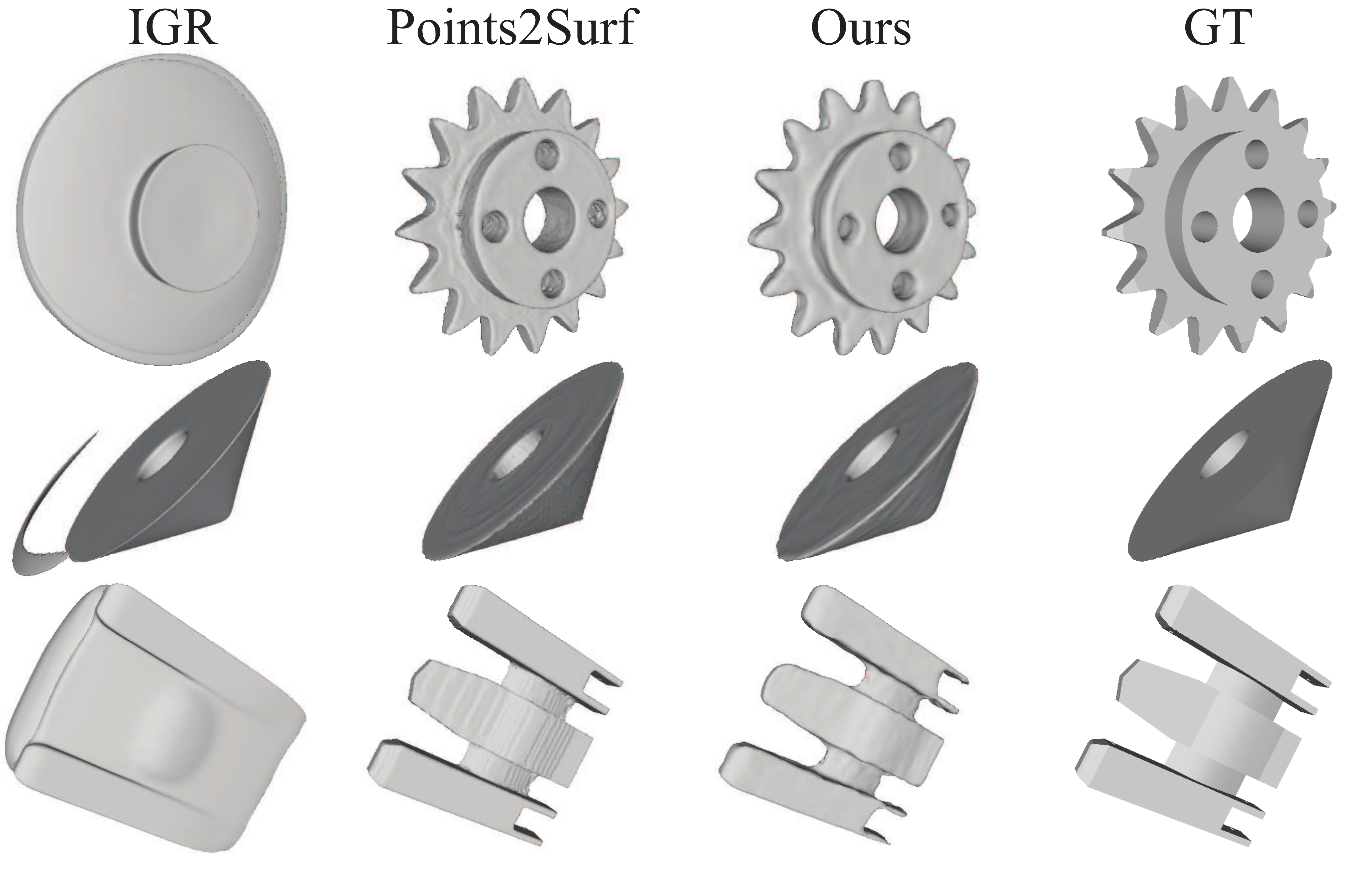}
  % replacing the above command with the one below will explicitly set
  % the bounding box of the PS figure to the rectangle (xl,yl),(xh,yh).
  % It will also prevent LaTeX from reading the PS file to determine
  % the bounding box (i.e., it will speed up the compilation process)
  % \includegraphics[width=.95\linewidth, bb=39 696 126 756]{sampleFig}
  %
  %
\caption{\label{fig:ABC} Comparison under ABC in surface reconstruction.}
\end{figure}

\begin{figure}[tb]
  \centering
  % the following command controls the width of the embedded PS file
  % (relative to the width of the current column)
  %\includegraphics[width=.95\linewidth, bb=39 696 126 756]{figures/definition3.eps}
   \includegraphics[width=\linewidth]{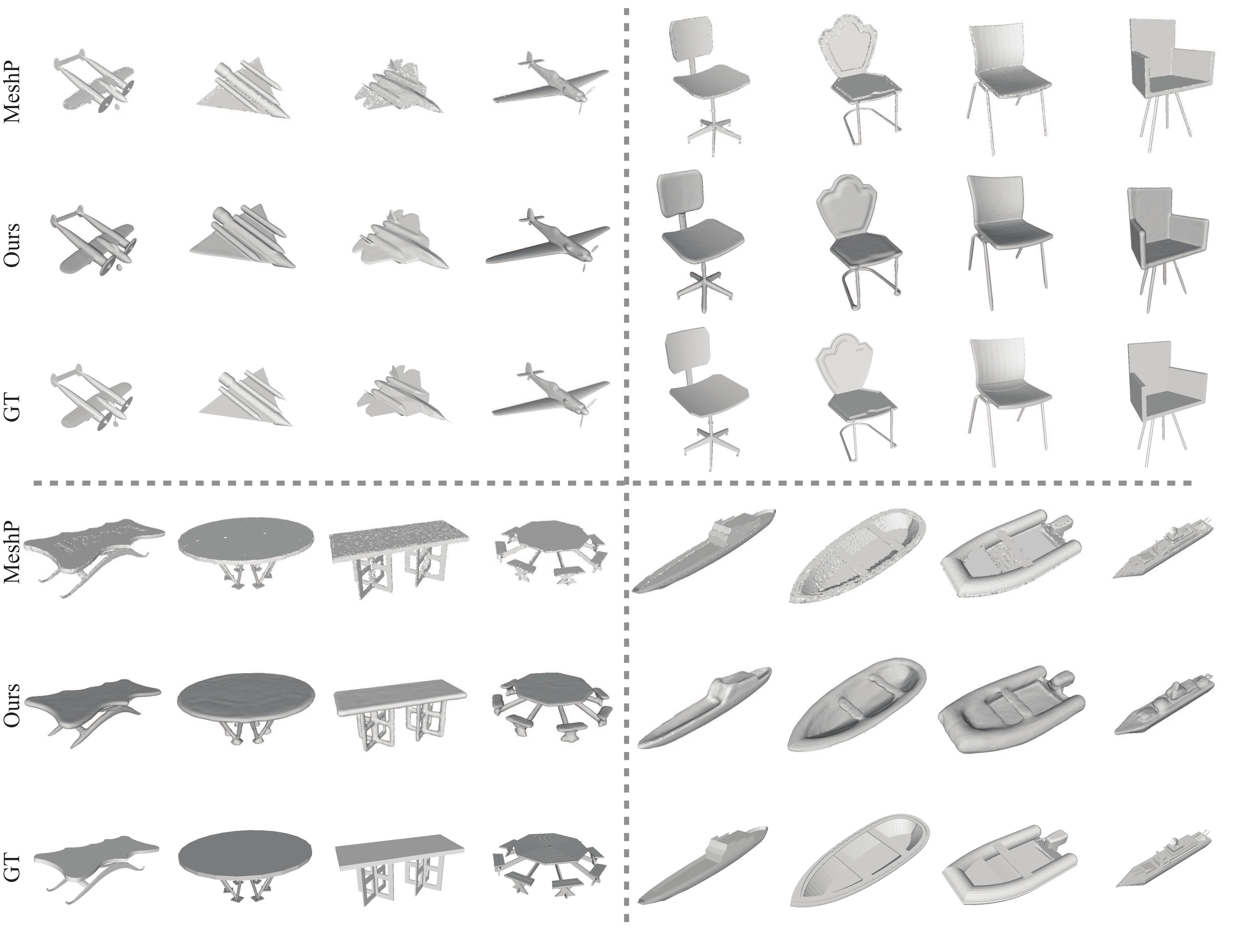}
  % replacing the above command with the one below will explicitly set
  % the bounding box of the PS figure to the rectangle (xl,yl),(xh,yh).
  % It will also prevent LaTeX from reading the PS file to determine
  % the bounding box (i.e., it will speed up the compilation process)
  % \includegraphics[width=.95\linewidth, bb=39 696 126 756]{sampleFig}
  %
  %
\caption{\label{fig:ShapeNet} Comparison under ShapeNet in surface reconstruction.}
\end{figure}

\noindent\textbf{Comparison. }We compare our method with state-of-the-art classic and data-driven surface reconstruction methods under the FAMOUS and ABC datasets, including DeepSDF (DSDF)~\cite{Park_2019_CVPR}, AtlasNet (ATLAS)~\cite{Groueix_2018_CVPR}, Screened Poisson Surface Reconstruction (PSR)~\cite{journals/tog/KazhdanH13}, Points2Surf~\cite{ErlerEtAl:Points2Surf:ECCV:2020}, and IGR~\cite{gropp2020implicit}. We report the results of DSDF, ATLAS, PSR and Points2Surf from the paper of Points2Surf~\cite{ErlerEtAl:Points2Surf:ECCV:2020}, while reproducing the results of IGR using the official code. The L2-CD comparison in Table~\ref{table:NOX1} shows that our method can significantly increase the surface reconstruction accuracy under each dataset due to better inference of the surface learned in the pulling process.

\begin{figure}[tb]
  \centering
  % the following command controls the width of the embedded PS file
  % (relative to the width of the current column)
  %\includegraphics[width=.95\linewidth, bb=39 696 126 756]{figures/definition3.eps}
   \includegraphics[width=\linewidth]{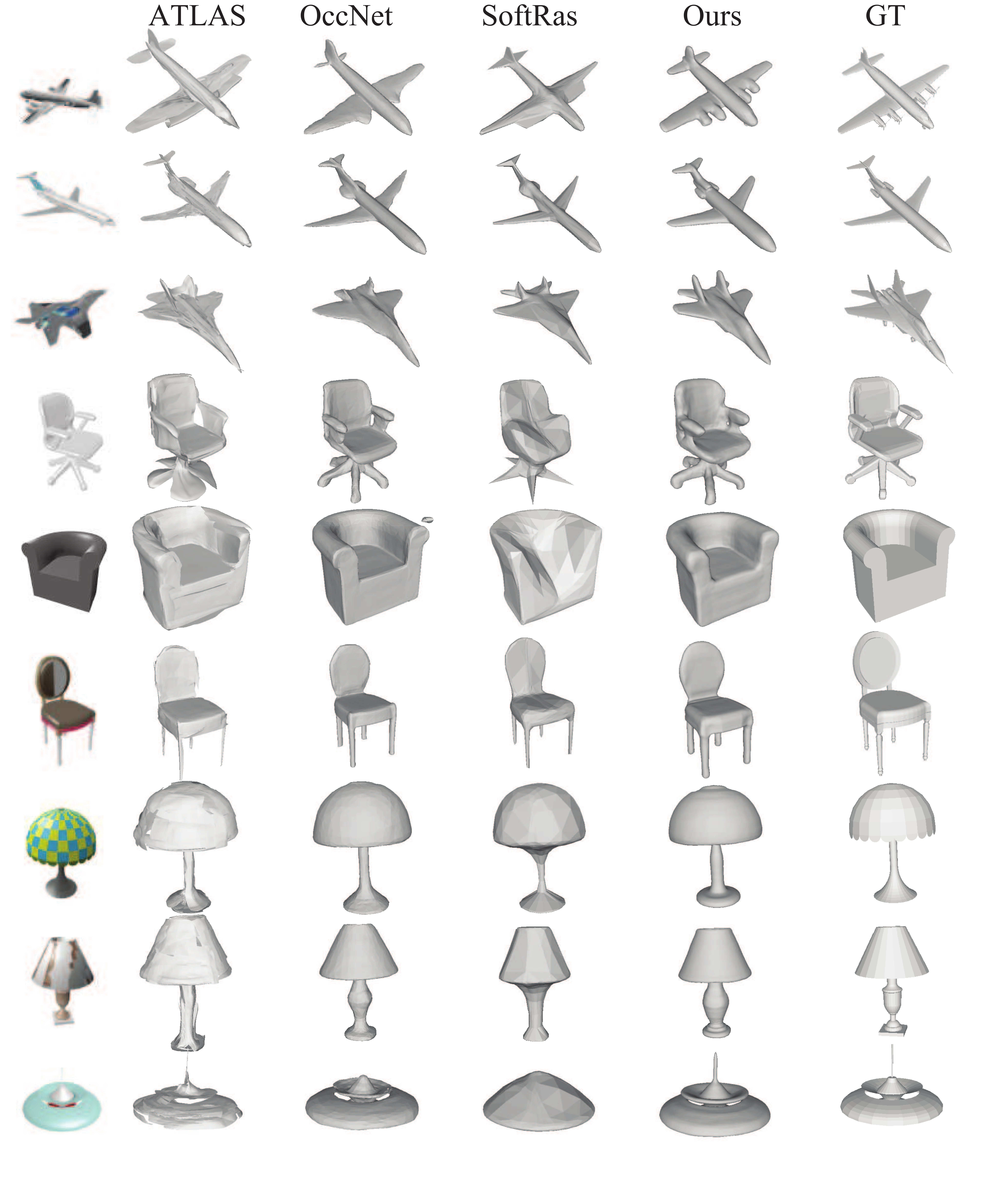}
  % replacing the above command with the one below will explicitly set
  % the bounding box of the PS figure to the rectangle (xl,yl),(xh,yh).
  % It will also prevent LaTeX from reading the PS file to determine
  % the bounding box (i.e., it will speed up the compilation process)
  % \includegraphics[width=.95\linewidth, bb=39 696 126 756]{sampleFig}
  %
  %
\caption{\label{fig:ShapeNetvis} Comparison in single image reconstruction.}
\end{figure}

We visually compare our method with IGR~\cite{gropp2020implicit} and Points2Surf~\cite{ErlerEtAl:Points2Surf:ECCV:2020} under the FAMOUS and ABC dataset in Fig.~\ref{fig:Famous} and Fig.~\ref{fig:ABC}, respectively. We train IGR using its released code with the same settings as ours, and generate surface reconstruction using the trained parameters released by Points2Surf. The comparison in Fig.~\ref{fig:Famous} demonstrates that our method can reveal geometry details in higher accuracy than other methods. Moreover, the comparison in Fig.~\ref{fig:ABC} shows that our method can reconstruct a smoother plane than Points2Surf, but Points2Surf is good at reconstructing sharp edges.

Similarly, we compare the state-of-the-art classic and data-driven methods under the ShapeNet subset, including PSR~\cite{journals/tog/KazhdanH13}, Ball-Pivoting algorithm (BPA)~\cite{817351TVCG}, ATLAS~\cite{Groueix_2018_CVPR}, Deep Geometric Prior (DGP)~\cite{Williams_2019_CVPR}, Deep Marching Cube (DMC)~\cite{Liao2018CVPR}, DeepSDF (DSDF)~\cite{Park_2019_CVPR}, MeshP~\cite{liu2020meshing}, Neural Unsigned Distance (NUD)~\cite{chibane2020neural}, SALD~\cite{atzmon2020sald}, Local SDF (GRID)~\cite{jiang2020lig}, and IMNET~\cite{chen2018implicit_decoder}. We conduct the numerical comparison in terms of different metrics including L2-CD in Table~\ref{table:t10}, normal consistency in Table~\ref{table:t11}, and F-score with a threshold of $\mu$ in Table~\ref{table:t12}, $2\mu$ in Table~\ref{table:t13}. We report the results of PSR, MC, BPA, ATLAS, DMC, DSDF, DGP, MeshP from the paper of MeshP~\cite{liu2020meshing}, while reporting the results of SALD, GRID, IMNET from their original papers and reproducing the results of NUD using the same experimental settings. The comparison shown in Table~\ref{table:t10},~\ref{table:t11},~\ref{table:t12},~\ref{table:t13} demonstrates that our method can reconstruct more accurate surfaces in terms of CD and F-Score, where we set the threshold $\mu$ as 0.002 in the F-Score calculation. Although our normal consistency results are comparable to MeshP, MeshP directly does the meshing without learning an implicit function, which requires dense and clean point clouds to guarantee the performance.

We visually compare our method with the-state-of-the-art MeshP~\cite{liu2020meshing} under Airplane, Chair, Table and Vessel classes in Fig.~\ref{fig:ShapeNet}. We use the parameters trained by MeshP. The comparison shows that our method can reconstruct more complete surfaces, especially for thin structures or sharp corners, which achieves much higher accuracy.

In addition, we also report our L1-CD results by comparing with 3D-R2~\cite{ChoyXGCS16}, PSGN~\cite{FanSG17}, DMC~\cite{Liao2018CVPR}, Occupancy Network (OccNet)~\cite{MeschederNetworks}, SSRNet~\cite{Mi_2020_CVPR} and DDT~\cite{luo2021deepdt} under the ShapeNet subset in Table~\ref{table:NOX2}. The comparison shows that our method achieves the best performance.

\begin{table}[tb]
\centering
\caption{Surface reconstruction comparison in terms of normal consistency.}  % ????????Comparison of shape reconstruction with known camera pose from silhouette images with different resolutions in terms of CD
\resizebox{\linewidth}{!}{
    \begin{tabular}{|c|c|c|c|c|c|c|c|c|c|c|c}  % ?????
     \hline
       %\cline{1-12}
       %\hline
        Class& PSR& DMC & BPA & ATLAS &DMC&DSDF&MeshP&GRID&IMNET&Ours \\  % ?????§á?
     \hline
        Display& 0.889& 0.842 & 0.952 & 0.828 &0.882&0.932& \textbf{0.974} & 0.926 &0.574&0.964\\
        Lamp &0.876&0.872&0.951&0.593&0.725&0.864&\textbf{0.963}&0.882&0.592&0.930\\
        Airplane&0.848&0.835&0.926&0.737&0.716&0.872&\textbf{0.955}&0.817&0.550&0.947\\
        Cabinet&0.880&0.827&0.836&0.682&0.845&0.872&\textbf{0.957}&0.948&0.700&0.930\\
        Vessel&0.861&0.831&0.917&0.671&0.706&0.841&\textbf{0.953}&0.847&0.574&0.941\\
        Table&0.833&0.809&0.919&0.783&0.831&0.901&\textbf{0.962}&0.936&0.702&0.908\\
        Chair&0.850&0.818&0.938&0.638&0.794&0.886&\textbf{0.962}&0.920&0.820&0.937\\
        Sofa&0.892&0.851&0.940&0.633&0.850&0.906&\textbf{0.971}&0.944&0.818&0.951\\
     \hline
     Mean&0.866&0.836&0.923&0.695&0.794&0.884&\textbf{0.962}&0.903&0.666&0.939\\
     \hline
   \end{tabular}}
   \label{table:t11}
\end{table}

\begin{table}[tb]
\centering
\caption{Surface reconstruction comparison in terms of F-score with a threshold of $\mu$.}  % ????????Comparison of shape reconstruction with known camera pose from silhouette images with different resolutions in terms of CD
\resizebox{\linewidth}{!}{
    \begin{tabular}{|c|c|c|c|c|c|c|c|c|c|c|c|c|c|c|c}  % ?????
     \hline
       %\cline{1-12}
       %\hline
        Class& PSR& DMC & BPA & ATLAS &DMC&DSDF& DGP &MeshP&NUD&GRID&IMNET&Ours \\  % ?????§á?
     \hline
        Display&0.468& 0.495& 0.834& 0.071& 0.108& 0.632& 0.417& 0.903& 0.903&0.551&0.601&\textbf{0.989}\\
        Lamp& 0.455 &0.518& 0.826& 0.029 &0.047& 0.268& 0.405& 0.855&0.888&0.624&0.836&\textbf{0.891}\\
        Airplane&0.415 &0.442& 0.788& 0.070& 0.050& 0.350& 0.249 &0.844&0.872&0.564&0.698&\textbf{0.996}\\
        Cabinet&0.392 &0.392& 0.553& 0.077& 0.154 &0.573& 0.513& 0.860&0.950&0.733&0.343&\textbf{0.980}\\
        Vessel&0.415& 0.466& 0.789& 0.058& 0.055& 0.323& 0.387& 0.862&0.883&0.467&0.147&\textbf{0.985}\\
        Table&0.233& 0.287& 0.772& 0.080 &0.095& 0.577& 0.307& 0.880& 0.908&0.844&0.425&\textbf{0.922}\\
        Chair&0.382 &0.433& 0.802& 0.050& 0.088& 0.447& 0.481& 0.875& 0.913&0.710&0.181&\textbf{0.954}\\
        Sofa&0.499& 0.535& 0.786& 0.058& 0.129& 0.577& 0.638& 0.895&0.945&0.822&0.199&\textbf{0.968}\\
     \hline
     Mean&0.407 &0.446 &0.769& 0.062 &0.091& 0.468& 0.425& 0.872&0.908&0.664&0.429&\textbf{0.961}\\
     \hline
   \end{tabular}}
   \label{table:t12}
\end{table}

\begin{table}[tb]
\centering
\caption{Surface reconstruction comparison in terms of F-score with a threshold of $2\mu$.}  % ????????Comparison of shape reconstruction with known camera pose from silhouette images with different resolutions in terms of CD
\resizebox{\linewidth}{!}{
    \begin{tabular}{|c|c|c|c|c|c|c|c|c|c|c|c|c|c|c}  % ?????
     \hline
       %\cline{1-12}
       %\hline
        Class& PSR& DMC & BPA & ATLAS &DMC&DSDF& DGP &MeshP&NUD&Ours \\  % ?????§á?
     \hline
        Display&0.666& 0.669& 0.929& 0.179& 0.246& 0.787& 0.607& 0.975&0.944&\textbf{0.991}\\
        Lamp&0.648& 0.681& 0.934& 0.077& 0.113& 0.478& 0.662& \textbf{0.951}&0.945&0.924\\
        Airplane&0.619& 0.639& 0.914& 0.179 &0.289& 0.566& 0.515& 0.946&0.944&\textbf{0.997}\\
        Cabinet&0.598& 0.591& 0.706& 0.195& 0.128& 0.694& 0.738& 0.946&0.980&\textbf{0.989}\\
        Vessel&0.633& 0.647& 0.906& 0.153& 0.120& 0.509& 0.648& 0.956&0.945&\textbf{0.990}\\
        Table&0.442& 0.462& 0.886& 0.195& 0.221& 0.743& 0.494& 0.963&0.922&\textbf{0.973}\\
        Chair&0.617& 0.615& 0.913& 0.134& 0.345& 0.665 & 0.693& 0.964&0.954&\textbf{0.969}\\
        Sofa&0.725& 0.708& 0.895& 0.153& 0.208& 0.734& 0.834& 0.972&0.968&\textbf{0.974}\\
     \hline
     Mean&0.618& 0.626& 0.885& 0.158& 0.209& 0.647& 0.649 &0.959&0.950&\textbf{0.976}\\
     \hline
   \end{tabular}}
   \label{table:t13}
\end{table}

\begin{table}[h]
\centering
\caption{Reconstruction comparison in terms of L1-CD.}  % ????????
\resizebox{\linewidth}{!}{
    \begin{tabular}{c|c|c|c|c|c|c}  % 32=0.1S£¬64=0.2S£¬128=1.3S * 600000
     \hline
          3D-R2 & PSGN & DMC& OccNet& SSRNet& DDT & Ours\\   % ?????§á?
     \hline
       0.169&0.202&0.117&0.079&0.024&0.020&\textbf{0.011} \\ %0.4806
     \hline
   \end{tabular}}
   \label{table:NOX2}
\end{table}

\subsection{Single Image Reconstruction}
\noindent\textbf{Details. }We further employ Neural-Pull to reconstruct 3D shapes from 2D images. We regard the 2D image as a condition, which corresponds to a 3D shape represented as a point cloud $\bm{P}$. During training, we leverage a condition and a set of query locations $\bm{Q}$ to minimize the loss in Eq.~\ref{eq:cd1}. During testing, we reconstruct a 3D shape from an input image with a given condition using marching cube~\cite{Lorensen87marchingcubes}. We leverage the 2D encoder used by SoftRas~\cite{liu2019softras} to infer the 2D image conditions.

\begin{table*}[tb]
\centering
\caption{Single image reconstruction comparison in terms of different metrics.}  % ????????Comparison of shape reconstruction with known camera pose from silhouette images with different resolutions in terms of CD
\resizebox{\linewidth}{!}{
    \begin{tabular}{c|cccccc|ccccc|cccccc}  % ?????
     \hline
        & \multicolumn{6}{|c|}{L1-CD,$10^5$ points}  & \multicolumn{5}{|c|}{Normal Consistency,$10^5$ points} & \multicolumn{6}{|c}{EMD$\times100$,2048 points} \\
     \hline
     & 3D-R2& PSGN & Pix2Mesh& ATLAS& OccNet &Ours& 3D-R2 & Pix2Mesh& ATLAS& OccNet &Ours& IMNET&3DN&Pix2Mesh& ATLAS& DISN&Ours\\
     Airplane & 0.227 &0.137&0.187&0.104&0.147&\textbf{0.016}& 0.629&0.759&0.836&0.840&\textbf{0.858}&2.90&3.30&2.98&3.39&2.67&\textbf{1.32}\\
     Bench & 0.194 &0.181&0.201&0.138&0.155&\textbf{0.016}& 0.678&0.732&0.779&0.813&\textbf{0.820}&2.80&2.98&2.58&3.22&2.48&\textbf{1.37}\\
     Cabinet& 0.217 &0.215&0.196&0.175&0.167&\textbf{0.018}& 0.782&0.834&0.850&0.879&\textbf{0.888}&3.14&3.21&3.44&3.36&3.04&\textbf{1.62}\\
     Car& 0.213 &0.169&0.180&0.141&0.159&\textbf{0.022}& 0.714&0.756&0.836&0.852&\textbf{0.861}&2.73&3.28&3.43&3.72&2.67&\textbf{1.56}\\
     Chair& 0.270 &0.247&0.265&0.209&0.228&\textbf{0.024}& 0.663&0.746&0.791&\textbf{0.823}&0.810&3.01&4.45&3.52&3.86&2.67&\textbf{2.03}\\
     Display& 0.314 &0.284&0.239&0.198&0.278&\textbf{0.020}& 0.720&0.830&0.858&0.854&\textbf{0.867}&2.81&3.91&2.92&3.12&2.73&\textbf{1.64}\\
     Lamp& 0.778 &0.314&0.308&0.305&0.479&\textbf{0.021}& 0.560&0.666&0.694&0.731&\textbf{0.867}&5.85&3.99&5.15&5.29&4.38&\textbf{2.85}\\
     Loudspeaker& 0.318&0.316&0.285&0.245&0.300&\textbf{0.032}& 0.711&0.782&0.825&0.832&\textbf{0.849}&3.80&4.47&3.56&3.75&3.47&\textbf{2.10}\\
     Rifle& 0.183&0.134&0.164&0.115&0.141&\textbf{0.019}& 0.670&0.718&0.725&0.766&\textbf{0.811}&2.65&2.78&3.04&3.35&2.30&\textbf{1.41}\\
     Sofa& 0.229&0.224&0.212&0.177&0.194&\textbf{0.019}& 0.731&0.820&0.840&\textbf{0.863}&0.856&2.71&3.31&2.70&3.14&2.62&\textbf{1.51}\\
     Table& 0.239&0.222&0.218&0.190&0.189&\textbf{0.025}& 0.732&0.784&0.832&\textbf{0.858}&0.810&3.39&3.94&3.52&3.98&3.11&\textbf{1.99}\\
     Telephone& 0.195&0.161&0.149&0.128&0.140&\textbf{0.018}& 0.817&0.907&0.923&0.935&\textbf{0.946}&2.14&2.70&2.66&3.19&2.06&\textbf{1.23}\\
     Vessel& 0.238&0.188&0.212&0.151&0.218&\textbf{0.027}& 0.629&0.699&0.756&0.794&\textbf{0.827}&2.75&3.92&3.94&4.39&2.77&\textbf{1.71}\\
     \hline
     Mean& 0.278&0.215&0.216&0.175&0.215&\textbf{0.021}& 0.695&0.772&0.811&0.834&\textbf{0.851}&3.13&3.56&3.34&3.67&2.84&\textbf{1.72}\\
     \hline
   \end{tabular}}
   \label{table:sir}
\end{table*}

\noindent\textbf{Dataset and Metric. }We use the ShapeNet subset released by Choy et al~\cite{ChoyXGCS16} to evaluate the performance in single image reconstruction, where the dataset also contains rendered RGB images in 13 shape classes and a train/test split. After getting the reconstructed meshes, we first leverage the L1-CD and Normal Consistency (NC) to evaluate the reconstruction error between the reconstructed shapes and the $1\times10^5$ ground truth points released by OccNet~\cite{MeschederNetworks}, where we uniformly sample $1\times10^5$ points on the reconstructed shapes. To evaluate our method in a multi-scale way, we also uniformly sample 2048 points on both of reconstructed shapes and $1\times10^5$ point ground truth to evaluate reconstruction error using Earth Mover Distance (EMD).

\noindent\textbf{Comparison. }We report numerical comparisons with 3D-R2~\cite{ChoyXGCS16}, PSGN~\cite{FanSG17}, Pix2Mesh~\cite{WangZLFLJ18}, ATLAS~\cite{Groueix_2018_CVPR}, OccNet~\cite{MeschederNetworks}, IMNET~\cite{chen2018implicit_decoder}, 3DN~\cite{wang20193dn}, DISN~\cite{xu2019disn} in Table~\ref{table:sir}. The comparison in terms of L1-CD and Normal Consistency shows that our method can significantly improve the reconstruction performance under almost all shape classes by providing more geometry details on the 3D shapes in higher resolution. The EMD comparison also shows our outperforming results over other methods under a sparse point setting. We further present a visual comparison with ATLAS~\cite{Groueix_2018_CVPR}, OccNet~\cite{MeschederNetworks} and SoftRas~\cite{liu2019softras} in Fig.~\ref{fig:ShapeNetvis} under Airplane, Chair and Lamp classes, which shows that we can reconstruct shapes with smoother surface in higher accuracy.

%\begin{table}[h]
%\centering
%\caption{Single image reconstruction comparison.}  % ????????
%\resizebox{\linewidth}{!}{
%    \begin{tabular}{c|c|c|c|c|c|c}  % 32=0.1S£¬64=0.2S£¬128=1.3S * 600000
%     \hline
%          &3D-R2 & PSGN & Pix2Mesh & AtlasNet & OccNet & Ours\\   % ?????§á?
%     \hline
%       \multirow{3}{*}{\rotatebox{90}{L1-CD}}&0.169&0.202&0.117&0.079&0.024&\textbf{0.011} \\ %0.4806
%     \hline
%   \end{tabular}}
%   \label{table:sir}
%\end{table}

\begin{figure}[tb]
  \centering
  % the following command controls the width of the embedded PS file
  % (relative to the width of the current column)
  %\includegraphics[width=.95\linewidth, bb=39 696 126 756]{figures/definition3.eps}
   \includegraphics[width=\linewidth]{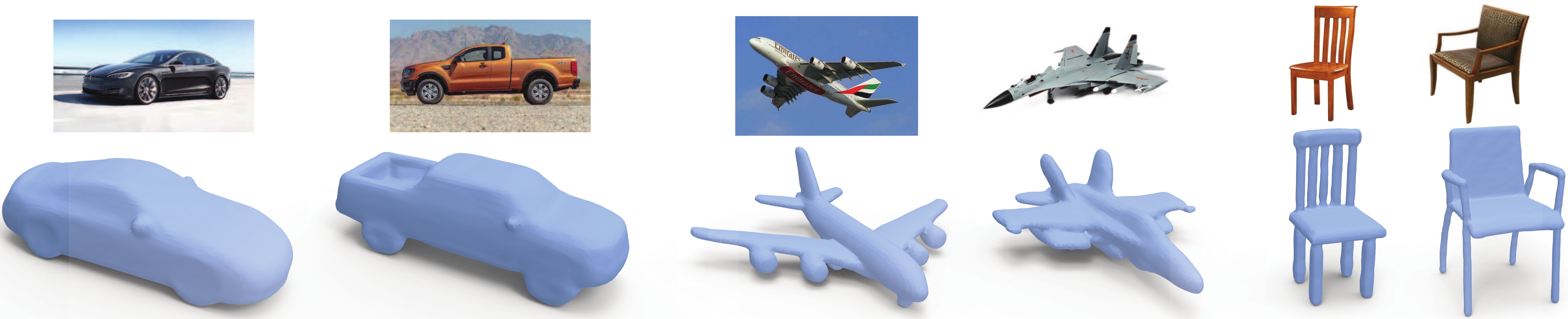}
  % replacing the above command with the one below will explicitly set
  % the bounding box of the PS figure to the rectangle (xl,yl),(xh,yh).
  % It will also prevent LaTeX from reading the PS file to determine
  % the bounding box (i.e., it will speed up the compilation process)
  % \includegraphics[width=.95\linewidth, bb=39 696 126 756]{sampleFig}
  %
  %
  %\vspace{-0.3in}
\caption{\label{fig:Teaser}Reconstruction from real images}
%\vspace{-0.3in}
\end{figure}

\noindent\textbf{Reconstruction from Real Images. }We collect some real images and reconstruct shapes using our model trained under synthetic data in Fig.~\ref{fig:Teaser}. The high fidelity reconstructions show that our method can generalize well to real images.

\subsection{Analysis}
\noindent\textbf{Ablation Study. }We conduct ablation studies in surface reconstruction under FAMOUS dataset. First, we explore the contribution made by the geometric network initialization (GNI). We report the result without GNI as ``No GNI'' using the random network initialization in Table~\ref{table:NOX8}. The degenerated result compared to our method of ``Ours'' demonstrates that GNI can help the network to better understand the shape.  Moreover, we highlight the strategy that we use in the query location sampling near the ground truth point clouds. We replace our sampling by randomly sampling query locations in the entire 3D space, where the number of query locations is kept the same. We report this result as ``Space sampling'' in Table~\ref{table:NOX8}, which demonstrates that it is more effective to use the query locations near the surface to probe the space for the learning. We also try to leverage an additional constraint introduced by IGR~\cite{gropp2020implicit} to keep the normal of gradient to be 1, but the result of ``Gradient constraint'' shows that the constraints bring no improvement.

\begin{table}[h]
\centering
\caption{Ablation studies in terms of L2-CD ($\times100$).}  % ????????
\resizebox{\linewidth}{!}{
    \begin{tabular}{c|c|c|c}  % 32=0.1S£¬64=0.2S£¬128=1.3S * 600000
     \hline
           No GNI & Space sampling & Gradient constraint& Ours\\   % ?????§á?
     \hline
      0.35& 0.80 & 1.15 & \textbf{0.22}\\ %0.4806
     \hline
   \end{tabular}}
   \label{table:NOX8}
\end{table}

\begin{figure}[tb]
  \centering
  % the following command controls the width of the embedded PS file
  % (relative to the width of the current column)
  %\includegraphics[width=.95\linewidth, bb=39 696 126 756]{figures/definition3.eps}
   \includegraphics[width=\linewidth]{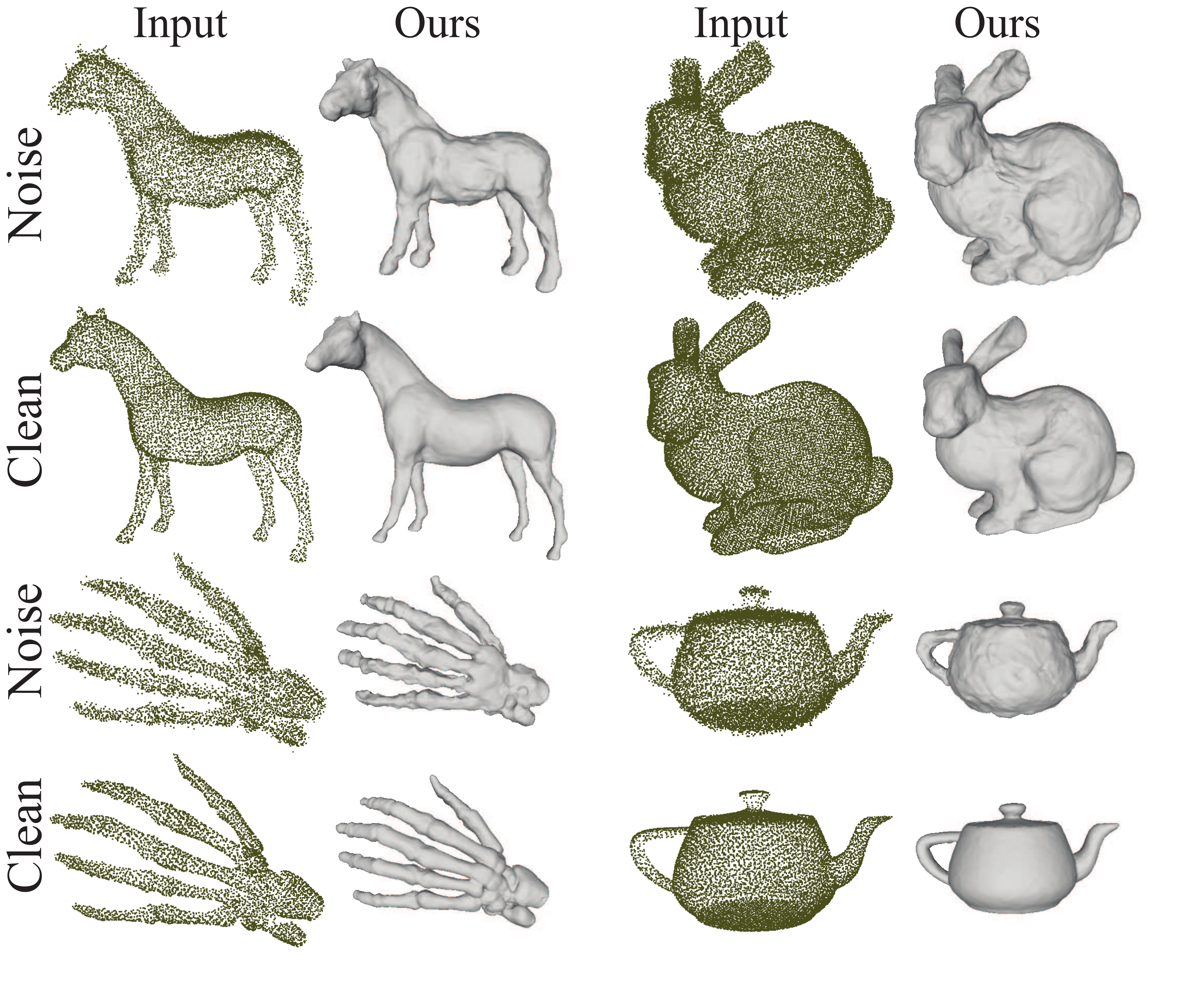}
  % replacing the above command with the one below will explicitly set
  % the bounding box of the PS figure to the rectangle (xl,yl),(xh,yh).
  % It will also prevent LaTeX from reading the PS file to determine
  % the bounding box (i.e., it will speed up the compilation process)
  % \includegraphics[width=.95\linewidth, bb=39 696 126 756]{sampleFig}
  %
  %
\caption{\label{fig:NoiseComp} Demonstration of resisting noise.}
\end{figure}

\noindent\textbf{The Effect of Noise. }We further explore the effect of noise on the ground truth point clouds under the ABC and FAMOUS datasets in surface reconstruction. We conduct experiments using the ``ABC max-noise'' and ``FAMOUS max-noise'' with strong noise, ``ABC var-noise'' with varying noise strength, and ``FAMOUS med-noise'' with a constant noise strength, all of which are released by Points2Surf~\cite{ErlerEtAl:Points2Surf:ECCV:2020}. We report our results under these datasets in Table~\ref{table:NOX3}, where we show that our method can better resist the noise than the state-of-the-art results. We also visually compare our results with noise and without noise under ``FAMOUS med-noise'' in Fig.~\ref{fig:NoiseComp}. The slight degeneration further demonstrates our ability of learning signed distance functions from point cloud with noise.

\begin{table}[h]
\centering
\caption{Comparison with noise in terms of L2-CD ($\times100$).}  % ????????
\resizebox{\linewidth}{!}{
    \begin{tabular}{c|c|c|c|c|c}  % 32=0.1S£¬64=0.2S£¬128=1.3S * 600000
     \hline
          Dataset& DSDF & ATLAS & PSR & Points2Surf& Ours\\   % ?????§á?
     \hline
       ABC var-noise& 12.51 & 4.04 & 3.29& 2.14&\textbf{0.72}  \\ %0.4806
       ABC max-noise& 11.34 & 4.47 & 3.89& 2.76&\textbf{1.24}  \\ %0.4806
       F-med-noise &9.89&4.54&1.80&1.51&\textbf{0.28}\\
       F-max-noise &13.17&4.14&3.41&2.52&\textbf{0.31}\\
     \hline
     Mean &11.73&4.30&3.10&2.23&\textbf{0.64}\\
     \hline
   \end{tabular}}
   \label{table:NOX3}
\end{table}

\noindent\textbf{The Effect of Query Location Resolution. }The number of query locations is also a factor that affects the learning. We explore its effect by merely adjusting the number of query locations under FAMOUS in surface reconstruction, such that $I=\{1,2.5,5,10\}\times10^6$. We report the comparison in Table~\ref{table:NOX4}, where the best result is achieved with $I=10\times10^6$. Moreover, we test the time used in training in one epoch for different numbers of query locations. Although the result with $I=10\times10^6$ is better than the one with $I=5\times10^6$ which is used in our previous experiments, it takes much more time in training.

\begin{table}[h]
\centering
\caption{Effect of $I$ in terms of L2-CD ($\times100$) and time.}  % ????????
%\resizebox{\linewidth}{!}{
    \begin{tabular}{c|c|c|c|c}  % 32=0.1S£¬64=0.2S£¬128=1.3S * 600000
     \hline
          $\times10^6$ & 1 & 2.5 & 5 & 10\\   % ?????§á?
     \hline
      Accuracy&0.434 & 0.394 &0.223 &\textbf{0.221}\\ %0.4806
      Time (s) & 103 & 210 & 530 &1020\\ %0.4806
     \hline
   \end{tabular}%}
   \label{table:NOX4}
\end{table}

\noindent\textbf{The Effect of GT Point Cloud Resolution. }We also explore how the resolution of ground truth point clouds affects the performance under the FAMOUS dataset in surface reconstruction in Table~\ref{table:NOX5}. We keep the number of query locations the same to $I=5\times10^6$, but employ ground truth point clouds with different numbers. Results in Table~\ref{table:NOX5} show that higher resolutions of the ground truth can help our method to better infer the surface, but it also takes much more time to search the nearest neighbor on the ground truth point cloud when calculating the loss, especially in real applications. Moreover, we also compare our method with DSDF, ATLAS, PSR, Points2Surf under FAMOUS sparser (``F-sparser'') and FAMOUS denser (``F-denser'') datasets released by Points2Surf~\cite{ErlerEtAl:Points2Surf:ECCV:2020}. Table~\ref{table:NOX1res} demonstrate that our method also achieves the best.

\begin{table}[h]
\centering
\caption{Effect of $J$ in terms of L2-CD ($\times100$).}  % ????????
%\resizebox{\linewidth}{!}{
    \begin{tabular}{c|c|c|c|c|c|c}  % 32=0.1S£¬64=0.2S£¬128=1.3S * 600000
     \hline
          $\times10^3$ & 1 & 2.5 & 5 & 10 & 20 &40\\   % ?????§á?
     \hline
          &0.293 & 0.266 &0.236 &0.233&0.223&\textbf{0.213}\\ %0.4806
     \hline
   \end{tabular}%}
   \label{table:NOX5}
\end{table}

\begin{table}[h]
\centering
\caption{Resolution comparison in terms of L2-CD ($\times100$).}  % ????????
\resizebox{\linewidth}{!}{
    \begin{tabular}{c|c|c|c|c|c}  % 32=0.1S£¬64=0.2S£¬128=1.3S * 600000
     \hline
          Dataset& DSDF & ATLAS & PSR & Points2Surf & Ours\\   % ?????§á?
     \hline
       F-sparse &10.41&4.91&2.17&1.93&\textbf{0.84}\\
       F-dense &9.49&4.35&1.60&1.33&\textbf{0.22}\\
     \hline
     Mean&9.60&4.66&1.98&1.62&\textbf{0.44}\\
     \hline
   \end{tabular}}
   \label{table:NOX1res}
\end{table}

\noindent\textbf{The Effect of Query Locations Range. }Finally, we discuss the effect of the query location range. We use the parameter $\sigma^2$ to control the maximum range of query locations around each point on the ground truth point cloud. We use several $\sigma^2$ candidates, including $\{0.25\sigma^2,0.5\sigma^2,\sigma^2,2\sigma^2,4\sigma^2\}$, to randomly sample the same number of query locations. We report the results under the FAMOUS dataset in surface reconstruction in Table~\ref{table:NOX6}. The comparison shows that a too small or too large query location range will degenerate the surface reconstruction performance. Since it is hard to use the query locations to probe the area around the surface if the query location range is too small, while it is also hard to push the network to produce the accurate direction and distance to move the query locations to the surface if the query locations are too far away from the surface.

\begin{table}[h]
\centering
\caption{Effect of $\sigma^2$ in terms of L2-CD ($\times100$).}  % ????????
%\resizebox{\linewidth}{!}{
    \begin{tabular}{c|c|c|c|c|c}  % 32=0.1S£¬64=0.2S£¬128=1.3S * 600000
     \hline
          $\times\sigma^2$ & 0.25& 0.5& 1 & 2 & 4\\   % ?????§á?
     \hline
          &0.348 & 0.304 &\textbf{0.223} &0.243&0.271\\ %0.4806
     \hline
   \end{tabular}%}
   \label{table:NOX6}
\end{table}

\begin{figure}[tb]
  \centering
  % the following command controls the width of the embedded PS file
  % (relative to the width of the current column)
  %\includegraphics[width=.95\linewidth, bb=39 696 126 756]{figures/definition3.eps}
   \includegraphics[width=\linewidth]{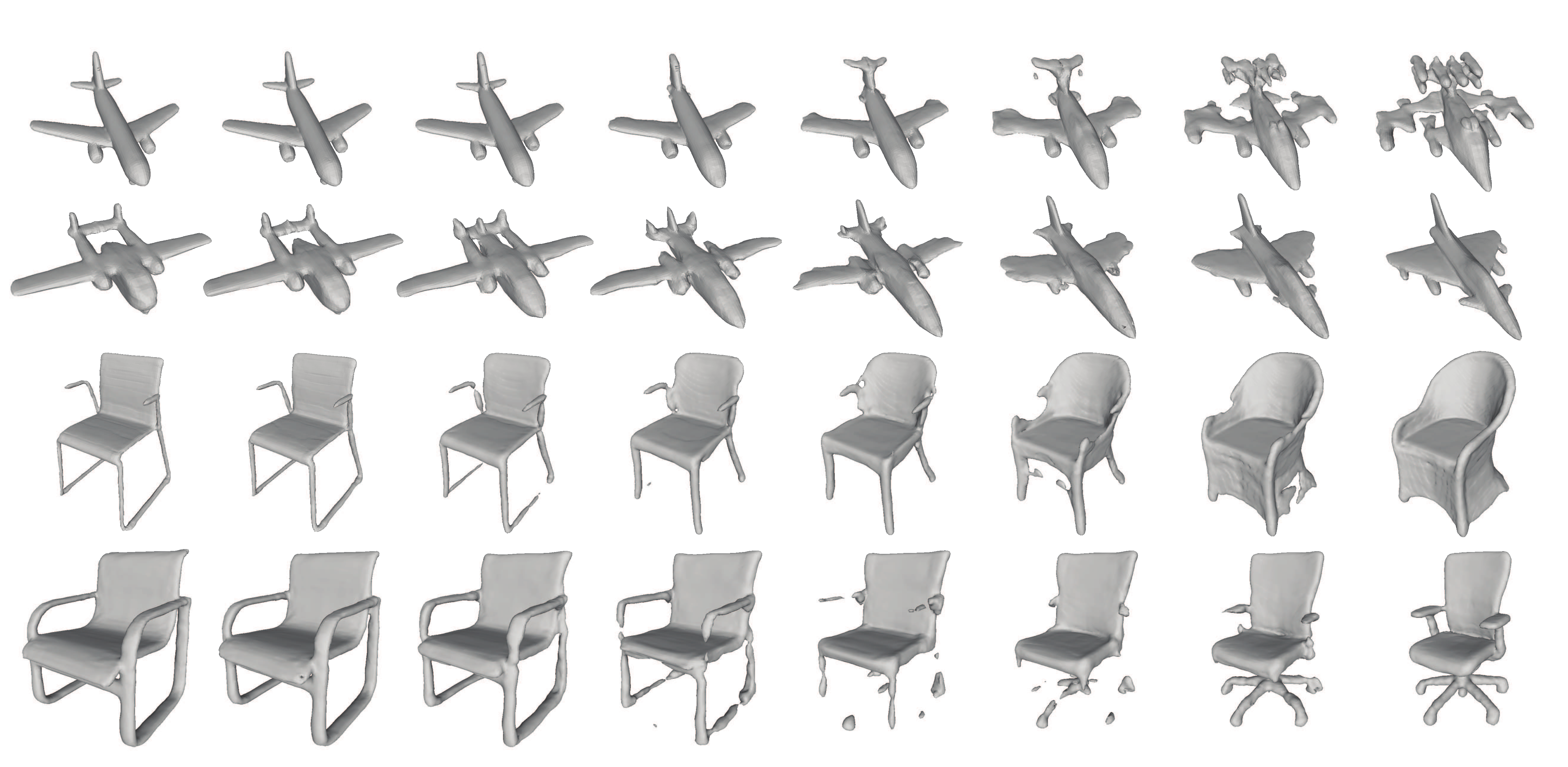}
  % replacing the above command with the one below will explicitly set
  % the bounding box of the PS figure to the rectangle (xl,yl),(xh,yh).
  % It will also prevent LaTeX from reading the PS file to determine
  % the bounding box (i.e., it will speed up the compilation process)
  % \includegraphics[width=.95\linewidth, bb=39 696 126 756]{sampleFig}
  %
  %
\caption{\label{fig:Iterpolation} Interpolated shapes in single image reconstruction.}
\end{figure}

\noindent\textbf{Latent Space Visualization. }We visualize the latent space learned by our network in single image reconstruction under ShapeNet subset. We randomly select two reconstructed shapes in the test set of Airplane class or Chair class, and regard their latent codes as two ends to interpolate six new latent codes between them. We leverage these interpolated latent codes to generate novel shapes by the trained point decoder. We visualize these shape interpolations under each one of Airplane and Chair classes in Fig.~\ref{fig:Iterpolation}, which shows that our method can reconstruct complex shapes with arbitrary topology. Moreover, the smooth transformation from one shape to another shape demonstrates that our method can help the network to learn a semantic latent space.

\begin{figure}[]
  \centering
  % the following command controls the width of the embedded PS file
  % (relative to the width of the current column)
  %\includegraphics[width=.95\linewidth, bb=39 696 126 756]{figures/definition3.eps}
   \includegraphics[width=\linewidth]{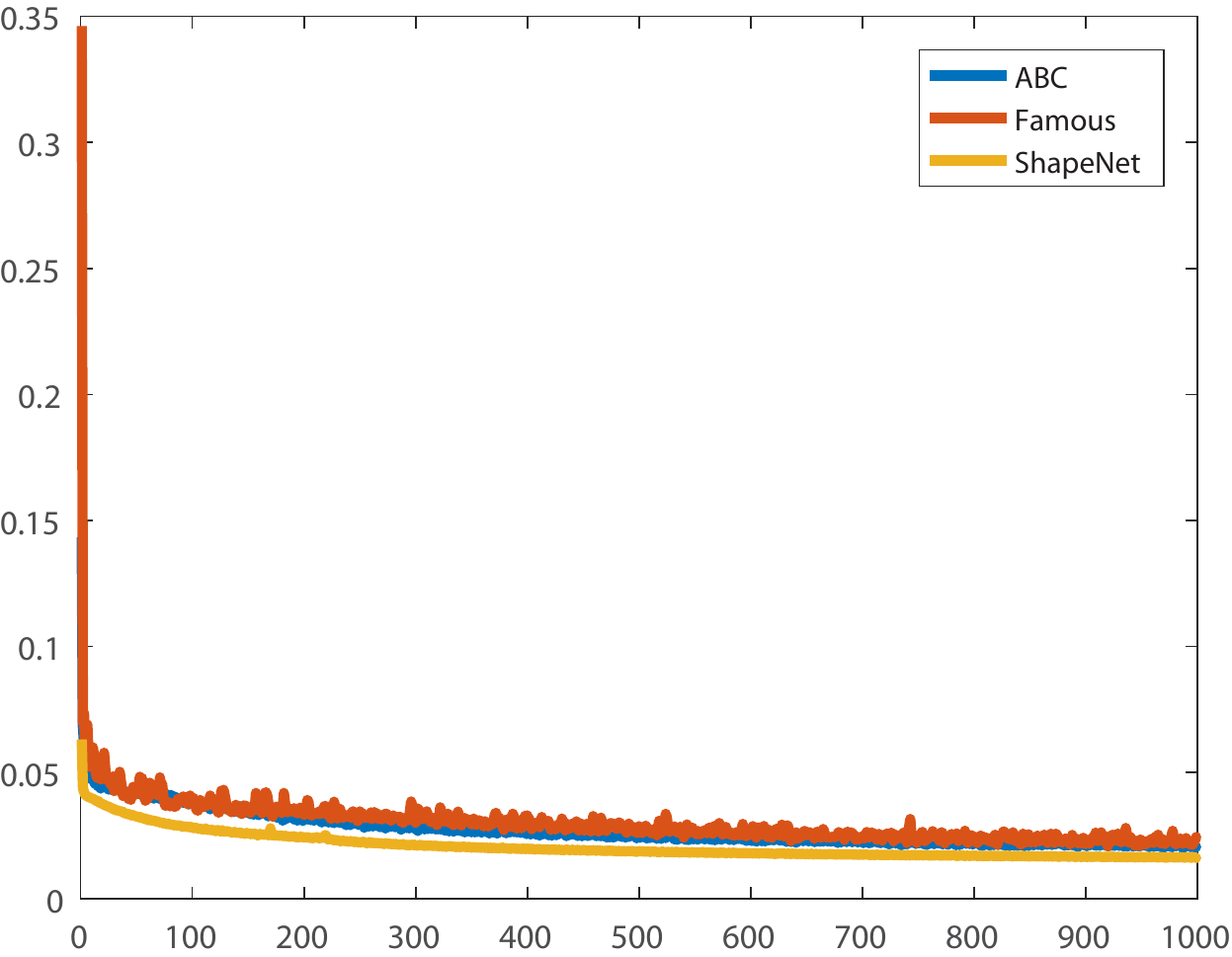}
  % replacing the above command with the one below will explicitly set
  % the bounding box of the PS figure to the rectangle (xl,yl),(xh,yh).
  % It will also prevent LaTeX from reading the PS file to determine
  % the bounding box (i.e., it will speed up the compilation process)
  % \includegraphics[width=.95\linewidth, bb=39 696 126 756]{sampleFig}
  %
  %
\caption{\label{fig:Comparison6} Training loss in surface reconstruction.}
\end{figure}

\noindent\textbf{Loss Visualization. }We further visualize the loss curves in surface reconstruction under ABC, Famous and ShapeNet dataset in Fig.~\ref{fig:Comparison6}. We can see that our method can effectively train a network to smoothly approach to the convergence.

\section{Conclusion}
We introduce Neural-Pull to learn signed distance functions from 3D point clouds by learning to pull 3D space onto the surface. Without the signed distance value ground truth, we can train a network to learn an SDF by pulling a sampled query location to its nearest neighbor on the surface. We effectively pull query locations along or against the gradient within the network with a stride of the predicted signed distance values. Being able to directly predict signed distances, our method successfully increases the 3D shape representation ability during testing. Our outperforming performance in single image reconstruction and surface reconstruction shows that we can reconstruct shapes and surfaces more accurately and flexibly than the state-of-the-art methods.

\section*{Acknowledgements}
We thank the anonymous reviewers for reviewing our paper and providing helpful comments. This work was supported by National Key R\&D Program of China (2020YFF0304100, 2018YFB0505400), the National Natural Science Foundation of China (62072268), and in part by Tsinghua-Kuaishou Institute of Future Media Data, and NSF (award 1813583).

\bibliography{paper}

\begin{thebibliography}{70}
\providecommand{\natexlab}[1]{#1}
\providecommand{\url}[1]{\texttt{#1}}
\expandafter\ifx\csname urlstyle\endcsname\relax
  \providecommand{\doi}[1]{doi: #1}\else
  \providecommand{\doi}{doi: \begingroup \urlstyle{rm}\Url}\fi

\bibitem[Atzmon \& Lipman(2020{\natexlab{a}})Atzmon and
  Lipman]{Atzmon_2020_CVPR}
Atzmon, M. and Lipman, Y.
\newblock Sal: Sign agnostic learning of shapes from raw data.
\newblock In \emph{IEEE Conference on Computer Vision and Pattern Recognition},
  2020{\natexlab{a}}.

\bibitem[Atzmon \& Lipman(2020{\natexlab{b}})Atzmon and Lipman]{atzmon2020sald}
Atzmon, M. and Lipman, Y.
\newblock {SALD}: Sign agnostic learning with derivatives.
\newblock \emph{arXiv}, 2006.05400, 2020{\natexlab{b}}.

\bibitem[Azinovic et~al.(2021)Azinovic, Martin{-}Brualla, Goldman, Nie{\ss}ner,
  and Thies]{DBLP:journals/corr/abs-2104-04532}
Azinovic, D., Martin{-}Brualla, R., Goldman, D.~B., Nie{\ss}ner, M., and Thies,
  J.
\newblock Neural {RGB-D} surface reconstruction.
\newblock \emph{CoRR}, abs/2104.04532, 2021.

\bibitem[Badki et~al.(2020)Badki, Gallo, Kautz, and Sen]{Badki_2020_CVPR}
Badki, A., Gallo, O., Kautz, J., and Sen, P.
\newblock Meshlet priors for {3D} mesh reconstruction.
\newblock In \emph{IEEE Conference on Computer Vision and Pattern Recognition},
  June 2020.

\bibitem[Bednarik et~al.(2020)Bednarik, Parashar, Gundogdu, and
  Salzmann]{bednarik2020}
Bednarik, J., Parashar, S., Gundogdu, E., and Salzmann, Mathieu~andFua, P.
\newblock Shape reconstruction by learning differentiable surface
  representations.
\newblock In \emph{{IEEE} Conference on Computer Vision and Pattern
  Recognition}, 2020.

\bibitem[{Bernardini} et~al.(1999){Bernardini}, {Mittleman}, {Rushmeier},
  {Silva}, and {Taubin}]{817351TVCG}
{Bernardini}, F., {Mittleman}, J., {Rushmeier}, H., {Silva}, C., and {Taubin},
  G.
\newblock The ball-pivoting algorithm for surface reconstruction.
\newblock \emph{IEEE Transactions on Visualization and Computer Graphics},
  5\penalty0 (4):\penalty0 349--359, 1999.

\bibitem[Chabra et~al.(2020)Chabra, Lenssen, Ilg, Schmidt, Straub, Lovegrove,
  and Newcombe]{DBLP:conf/eccv/ChabraLISSLN20}
Chabra, R., Lenssen, J.~E., Ilg, E., Schmidt, T., Straub, J., Lovegrove, S.,
  and Newcombe, R.~A.
\newblock Deep local shapes: Learning local {SDF} priors for detailed {3D}
  reconstruction.
\newblock In Vedaldi, A., Bischof, H., Brox, T., and Frahm, J. (eds.),
  \emph{European Conference on Computer Vision}, volume 12374, pp.\  608--625,
  2020.

\bibitem[Chang et~al.(2015)Chang, Funkhouser, Guibas, Hanrahan, Huang, Li,
  Savarese, Savva, Song, Su, Xiao, Yi, and Yu]{shapenet2015}
Chang, A.~X., Funkhouser, T., Guibas, L., Hanrahan, P., Huang, Q., Li, Z.,
  Savarese, S., Savva, M., Song, S., Su, H., Xiao, J., Yi, L., and Yu, F.
\newblock {Shape{N}et: An Information-Rich 3D Model Repository}.
\newblock Technical Report arXiv:1512.03012 [cs.GR], Stanford University ---
  Princeton University --- Toyota Technological Institute at Chicago, 2015.

\bibitem[Chen \& Zhang(2019)Chen and Zhang]{chen2018implicit_decoder}
Chen, Z. and Zhang, H.
\newblock Learning implicit fields for generative shape modeling.
\newblock \emph{IEEE Conference on Computer Vision and Pattern Recognition},
  2019.

\bibitem[Chibane et~al.(2020{\natexlab{a}})Chibane, Alldieck, and
  Pons{-}Moll]{DBLP:conf/cvpr/ChibaneAP20}
Chibane, J., Alldieck, T., and Pons{-}Moll, G.
\newblock Implicit functions in feature space for 3d shape reconstruction and
  completion.
\newblock In \emph{IEEE Conference on Computer Vision and Pattern Recognition},
  pp.\  6968--6979, 2020{\natexlab{a}}.

\bibitem[Chibane et~al.(2020{\natexlab{b}})Chibane, Mir, and
  Pons-Moll]{chibane2020neural}
Chibane, J., Mir, A., and Pons-Moll, G.
\newblock Neural unsigned distance fields for implicit function learning.
\newblock \emph{arXiv}, 2010.13938, 2020{\natexlab{b}}.

\bibitem[Choy et~al.(2016)Choy, Xu, Gwak, Chen, and Savarese]{ChoyXGCS16}
Choy, C.~B., Xu, D., Gwak, J., Chen, K., and Savarese, S.
\newblock {3D-R2N2}: {A} unified approach for single and multi-view 3{D} object
  reconstruction.
\newblock In \emph{European Conference on Computer Vision}, pp.\  628--644,
  2016.

\bibitem[Dupont et~al.(2021)Dupont, Teh, and
  Doucet]{DBLP:journals/corr/abs-2102-04776}
Dupont, E., Teh, Y.~W., and Doucet, A.
\newblock Generative models as distributions of functions.
\newblock \emph{CoRR}, abs/2102.04776, 2021.

\bibitem[Erler et~al.(2020)Erler, Guerrero, Ohrhallinger, Mitra, and
  Wimmer]{ErlerEtAl:Points2Surf:ECCV:2020}
Erler, P., Guerrero, P., Ohrhallinger, S., Mitra, N.~J., and Wimmer, M.
\newblock {Points2Surf}: Learning implicit surfaces from point clouds.
\newblock In Vedaldi, A., Bischof, H., Brox, T., and Frahm, J.-M. (eds.),
  \emph{European Conference on Computer Vision}, 2020.

\bibitem[Fan et~al.(2017)Fan, Su, and Guibas]{FanSG17}
Fan, H., Su, H., and Guibas, L.~J.
\newblock A point set generation network for 3{D} object reconstruction from a
  single image.
\newblock In \emph{{IEEE} Conference on Computer Vision and Pattern
  Recognition}, pp.\  2463--2471, 2017.

\bibitem[Genova et~al.(2019)Genova, Cole, Vlasic, Sarna, Freeman, and
  Funkhouser]{Genova:2019:LST}
Genova, K., Cole, F., Vlasic, D., Sarna, A., Freeman, W.~T., and Funkhouser, T.
\newblock Learning shape templates with structured implicit functions.
\newblock In \emph{International Conference on Computer Vision}, 2019.

\bibitem[Genova et~al.(2020)Genova, Cole, Sud, Sarna, and
  Funkhouser]{Genova_2020_CVPR}
Genova, K., Cole, F., Sud, A., Sarna, A., and Funkhouser, T.
\newblock Local deep implicit functions for 3d shape.
\newblock In \emph{IEEE Conference on Computer Vision and Pattern Recognition},
  June 2020.

\bibitem[Gropp et~al.(2020)Gropp, Yariv, Haim, Atzmon, and
  Lipman]{gropp2020implicit}
Gropp, A., Yariv, L., Haim, N., Atzmon, M., and Lipman, Y.
\newblock Implicit geometric regularization for learning shapes.
\newblock \emph{arXiv}, 2002.10099, 2020.

\bibitem[Groueix et~al.(2018)Groueix, Fisher, Kim, Russell, and
  Aubry]{Groueix_2018_CVPR}
Groueix, T., Fisher, M., Kim, V.~G., Russell, B.~C., and Aubry, M.
\newblock A papier-mâché approach to learning 3{D} surface generation.
\newblock In \emph{IEEE Conference on Computer Vision and Pattern Recognition},
  2018.

\bibitem[Han et~al.(2019{\natexlab{a}})Han, Shang, Liu, and
  Zwicker]{Zhizhong2018VIP}
Han, Z., Shang, M., Liu, Y.-S., and Zwicker, M.
\newblock {View Inter-Prediction GAN}: Unsupervised representation learning for
  3{D} shapes by learning global shape memories to support local view
  predictions.
\newblock In \emph{AAAI}, pp.\  8376--8384, 2019{\natexlab{a}}.

\bibitem[Han et~al.(2019{\natexlab{b}})Han, Shang, Liu, Vong, Liu, Zwicker,
  Han, and Chen]{Zhizhong2018seq}
Han, Z., Shang, M., Liu, Z., Vong, C.-M., Liu, Y.-S., Zwicker, M., Han, J., and
  Chen, C.~P.
\newblock {SeqViews2SeqLabels}: Learning 3{D} global features via aggregating
  sequential views by rnn with attention.
\newblock \emph{IEEE Transactions on Image Processing}, 28\penalty0
  (2):\penalty0 685--672, 2019{\natexlab{b}}.

\bibitem[Han et~al.(2019{\natexlab{c}})Han, Wang, Liu, and Zwicker]{MAPVAE19}
Han, Z., Wang, X., Liu, Y.-S., and Zwicker, M.
\newblock Multi-angle point cloud-vae:unsupervised feature learning for 3{D}
  point clouds from multiple angles by joint self-reconstruction and
  half-to-half prediction.
\newblock In \emph{{IEEE} International Conference on Computer Vision},
  2019{\natexlab{c}}.

\bibitem[Han et~al.(2019{\natexlab{d}})Han, Wang, Vong, Liu, Zwicker, and
  Chen]{3DViewGraph19}
Han, Z., Wang, X., Vong, C.-M., Liu, Y.-S., Zwicker, M., and Chen, C.~P.
\newblock {3{D}ViewGraph}: Learning global features for 3{D} shapes from a
  graph of unordered views with attention.
\newblock In \emph{IJCAI}, 2019{\natexlab{d}}.

\bibitem[Han et~al.(2020{\natexlab{a}})Han, Chen, Liu, and
  Zwicker]{Han2019ShapeCaptionerGCacmmm}
Han, Z., Chen, C., Liu, Y.-S., and Zwicker, M.
\newblock {ShapeCaptioner}: Generative caption network for {3D} shapes by
  learning a mapping from parts detected in multiple views to sentences.
\newblock In \emph{ACM International Conference on Multimedia},
  2020{\natexlab{a}}.

\bibitem[Han et~al.(2020{\natexlab{b}})Han, Chen, Liu, and
  Zwicker]{handrwr2020}
Han, Z., Chen, C., Liu, Y.-S., and Zwicker, M.
\newblock {DRWR}: A differentiable renderer without rendering for unsupervised
  3{D} structure learning from silhouette images.
\newblock In \emph{International Conference on Machine Learning},
  2020{\natexlab{b}}.

\bibitem[Han et~al.(2020{\natexlab{c}})Han, Qiao, Liu, and
  Zwicker]{seqxy2seqzeccv2020paper}
Han, Z., Qiao, G., Liu, Y.-S., and Zwicker, M.
\newblock {SeqXY2SeqZ}: Structure learning for {3D} shapes by sequentially
  predicting {1D} occupancy segments from {2D} coordinates.
\newblock In \emph{European Conference on Computer Vision}, 2020{\natexlab{c}}.

\bibitem[Hu et~al.(2020)Hu, Han, and Zwicker]{hutaoaaai2020}
Hu, T., Han, Z., and Zwicker, M.
\newblock 3{D} shape completion with multi-view consistent inference.
\newblock In \emph{AAAI}, 2020.

\bibitem[Jia \& Kyan(2020)Jia and Kyan]{jia2020learning}
Jia, M. and Kyan, M.
\newblock Learning occupancy function from point clouds for surface
  reconstruction.
\newblock \emph{arXiv}, 2010.11378, 2020.

\bibitem[Jiang et~al.(2020{\natexlab{a}})Jiang, Sud, Makadia, Huang,
  Nie{\ss}ner, and Funkhouser]{jiang2020lig}
Jiang, C., Sud, A., Makadia, A., Huang, J., Nie{\ss}ner, M., and Funkhouser, T.
\newblock Local implicit grid representations for {3D} scenes.
\newblock In \emph{IEEE Conference on Computer Vision and Pattern Recognition},
  2020{\natexlab{a}}.

\bibitem[Jiang et~al.(2020{\natexlab{b}})Jiang, Ji, Han, and
  Zwicker]{Jiang2019SDFDiffDRcvpr}
Jiang, Y., Ji, D., Han, Z., and Zwicker, M.
\newblock {SDFDiff}: Differentiable rendering of signed distance fields for
  {3D} shape optimization.
\newblock In \emph{IEEE Conference on Computer Vision and Pattern Recognition},
  2020{\natexlab{b}}.

\bibitem[Kazhdan \& Hoppe(2013)Kazhdan and Hoppe]{journals/tog/KazhdanH13}
Kazhdan, M.~M. and Hoppe, H.
\newblock Screened poisson surface reconstruction.
\newblock \emph{ACM Transactions Graphics}, 32\penalty0 (3):\penalty0
  29:1--29:13, 2013.

\bibitem[Koch et~al.(2019)Koch, Matveev, Jiang, Williams, Artemov, Burnaev,
  Alexa, Zorin, and Panozzo]{Koch_2019_CVPR}
Koch, S., Matveev, A., Jiang, Z., Williams, F., Artemov, A., Burnaev, E.,
  Alexa, M., Zorin, D., and Panozzo, D.
\newblock {ABC}: A big cad model dataset for geometric deep learning.
\newblock In \emph{IEEE Conference on Computer Vision and Pattern Recognition},
  June 2019.

\bibitem[Liao et~al.(2018)Liao, Donn\'{e}, and Geiger]{Liao2018CVPR}
Liao, Y., Donn\'{e}, S., and Geiger, A.
\newblock Deep marching cubes: Learning explicit surface representations.
\newblock In \emph{Conference on Computer Vision and Pattern Recognition},
  2018.

\bibitem[Littwin \& Wolf(2019)Littwin and Wolf]{Gidi_2019_ICCV}
Littwin, G. and Wolf, L.
\newblock Deep meta functionals for shape representation.
\newblock In \emph{{IEEE} International Conference on Computer Vision}, 2019.

\bibitem[Liu et~al.()Liu, Zhang, and Su]{liu2020meshing}
Liu, M., Zhang, X., and Su, H.
\newblock Meshing point clouds with predicted intrinsic-extrinsic ratio
  guidance.
\newblock In \emph{European Conference on Computer vision}.

\bibitem[Liu et~al.(2019{\natexlab{a}})Liu, Li, Chen, and Li]{liu2019softras}
Liu, S., Li, T., Chen, W., and Li, H.
\newblock Soft rasterizer: A differentiable renderer for image-based 3{D}
  reasoning.
\newblock \emph{IEEE International Conference on Computer Vision},
  2019{\natexlab{a}}.

\bibitem[Liu et~al.(2019{\natexlab{b}})Liu, Saito, Chen, and Li]{shichenNIPS}
Liu, S., Saito, S., Chen, W., and Li, H.
\newblock Learning to infer implicit surfaces without 3{D} supervision.
\newblock In \emph{Advances in Neural Information Processing Systems},
  2019{\natexlab{b}}.

\bibitem[Liu et~al.(2020)Liu, Zhang, Peng, Shi, Pollefeys, and
  Cui]{DIST2019SDFRcvpr}
Liu, S., Zhang, Y., Peng, S., Shi, B., Pollefeys, M., and Cui, Z.
\newblock {DIST}: Rendering deep implicit signed distance function with
  differentiable sphere tracing.
\newblock In \emph{IEEE Conference on Computer Vision and Pattern Recognition},
  2020.

\bibitem[Liu et~al.(2019{\natexlab{c}})Liu, Han, Liu, and Zwicker]{p2seq18}
Liu, X., Han, Z., Liu, Y.-S., and Zwicker, M.
\newblock {Point2Sequence}: Learning the shape representation of 3{D} point
  clouds with an attention-based sequence to sequence network.
\newblock In \emph{AAAI}, pp.\  8778--8785, 2019{\natexlab{c}}.

\bibitem[Liu et~al.(2021)Liu, Han, Liu, and Zwicker]{9318534}
Liu, X., Han, Z., Liu, Y.-S., and Zwicker, M.
\newblock Fine-grained 3d shape classification with hierarchical part-view
  attention.
\newblock \emph{IEEE Transactions on Image Processing}, 30:\penalty0
  1744--1758, 2021.

\bibitem[Lorensen \& Cline(1987)Lorensen and Cline]{Lorensen87marchingcubes}
Lorensen, W.~E. and Cline, H.~E.
\newblock Marching cubes: A high resolution 3d surface construction algorithm.
\newblock \emph{Computer Graphics}, 21\penalty0 (4):\penalty0 163--169, 1987.

\bibitem[Luo et~al.(2020)Luo, Mi, and Tao]{luo2021deepdt}
Luo, Y., Mi, Z., and Tao, W.
\newblock Deepdt: Learning geometry from delaunay triangulation for surface
  reconstruction.
\newblock \emph{CoRR}, abs/2101.10353, 2020.

\bibitem[Martel et~al.(2021)Martel, Lindell, Lin, Chan, Monteiro, and
  Wetzstein]{DBLP:journals/corr/abs-2105-02788}
Martel, J. N.~P., Lindell, D.~B., Lin, C.~Z., Chan, E.~R., Monteiro, M., and
  Wetzstein, G.
\newblock {ACORN:} adaptive coordinate networks for neural scene
  representation.
\newblock \emph{CoRR}, abs/2105.02788, 2021.

\bibitem[Mescheder et~al.(2019)Mescheder, Oechsle, Niemeyer, Nowozin, and
  Geiger]{MeschederNetworks}
Mescheder, L., Oechsle, M., Niemeyer, M., Nowozin, S., and Geiger, A.
\newblock Occupancy networks: Learning 3{D} reconstruction in function space.
\newblock In \emph{IEEE Conference on Computer Vision and Pattern Recognition},
  2019.

\bibitem[Mi et~al.(2020)Mi, Luo, and Tao]{Mi_2020_CVPR}
Mi, Z., Luo, Y., and Tao, W.
\newblock Ssrnet: Scalable 3{D} surface reconstruction network.
\newblock In \emph{IEEE Conference on Computer Vision and Pattern Recognition},
  June 2020.

\bibitem[Michalkiewicz et~al.(2019)Michalkiewicz, Pontes, Jack, Baktashmotlagh,
  and Eriksson]{DBLP:journals/corr/abs-1901-06802}
Michalkiewicz, M., Pontes, J.~K., Jack, D., Baktashmotlagh, M., and Eriksson,
  A.~P.
\newblock Deep level sets: Implicit surface representations for 3{D} shape
  inference.
\newblock \emph{CoRR}, abs/1901.06802, 2019.

\bibitem[Mildenhall et~al.(2020)Mildenhall, Srinivasan, Tancik, Barron,
  Ramamoorthi, and Ng]{mildenhall2020nerf}
Mildenhall, B., Srinivasan, P.~P., Tancik, M., Barron, J.~T., Ramamoorthi, R.,
  and Ng, R.
\newblock Nerf: Representing scenes as neural radiance fields for view
  synthesis.
\newblock In \emph{European Conference on Computer Vision}, 2020.

\bibitem[Niemeyer et~al.(2020)Niemeyer, Mescheder, Oechsle, and
  Geiger]{Volumetric2019SDFRcvpr}
Niemeyer, M., Mescheder, L., Oechsle, M., and Geiger, A.
\newblock Differentiable volumetric rendering: Learning implicit 3{D}
  representations without 3{D} supervision.
\newblock In \emph{IEEE Conference on Computer Vision and Pattern Recognition},
  2020.

\bibitem[Oechsle et~al.(2021)Oechsle, Peng, and
  Geiger]{DBLP:journals/corr/abs-2104-10078}
Oechsle, M., Peng, S., and Geiger, A.
\newblock {UNISURF:} unifying neural implicit surfaces and radiance fields for
  multi-view reconstruction.
\newblock \emph{CoRR}, abs/2104.10078, 2021.

\bibitem[Ost et~al.(2020)Ost, Mannan, Thuerey, Knodt, and
  Heide]{DBLP:journals/corr/abs-2011-10379}
Ost, J., Mannan, F., Thuerey, N., Knodt, J., and Heide, F.
\newblock Neural scene graphs for dynamic scenes.
\newblock \emph{CoRR}, abs/2011.10379, 2020.

\bibitem[Park et~al.(2019)Park, Florence, Straub, Newcombe, and
  Lovegrove]{Park_2019_CVPR}
Park, J.~J., Florence, P., Straub, J., Newcombe, R., and Lovegrove, S.
\newblock {DeepSDF}: Learning continuous signed distance functions for shape
  representation.
\newblock In \emph{IEEE Conference on Computer Vision and Pattern Recognition},
  2019.

\bibitem[Rematas et~al.(2021)Rematas, Martin-Brualla, and
  Ferrari]{rematasICML21}
Rematas, K., Martin-Brualla, R., and Ferrari, V.
\newblock Sharf: Shape-conditioned radiance fields from a single view.
\newblock In \emph{ICML}, 2021.

\bibitem[Saito et~al.(2019)Saito, , Huang, Natsume, Morishima, Kanazawa, and
  Li]{pifuSHNMKL19}
Saito, S., , Huang, Z., Natsume, R., Morishima, S., Kanazawa, A., and Li, H.
\newblock {PIFu}: Pixel-aligned implicit function for high-resolution clothed
  human digitization.
\newblock \emph{{IEEE} International Conference on Computer Vision}, 2019.

\bibitem[Sitzmann et~al.(2019)Sitzmann, Zollh{\"o}fer, and
  Wetzstein]{sitzmann2019srns}
Sitzmann, V., Zollh{\"o}fer, M., and Wetzstein, G.
\newblock Scene representation networks: Continuous 3{D}-structure-aware neural
  scene representations.
\newblock In \emph{Advances in Neural Information Processing Systems}, 2019.

\bibitem[Sitzmann et~al.(2020)Sitzmann, Martel, Bergman, Lindell, and
  Wetzstein]{sitzmann2019siren}
Sitzmann, V., Martel, J.~N., Bergman, A.~W., Lindell, D.~B., and Wetzstein, G.
\newblock Implicit neural representations with periodic activation functions.
\newblock In \emph{NeurIPS}, 2020.

\bibitem[Songyou~Peng(2020)]{Peng2020ECCV}
Songyou~Peng, Michael~Niemeyer, L. M. M. P. A.~G.
\newblock Convolutional occupancy networks.
\newblock In \emph{European Conference on Computer Vision}, 2020.

\bibitem[Takikawa et~al.(2021)Takikawa, Litalien, Yin, Kreis, Loop,
  Nowrouzezahrai, Jacobson, McGuire, and Fidler]{takikawa2021nglod}
Takikawa, T., Litalien, J., Yin, K., Kreis, K., Loop, C., Nowrouzezahrai, D.,
  Jacobson, A., McGuire, M., and Fidler, S.
\newblock Neural geometric level of detail: Real-time rendering with implicit
  {3D} shapes.
\newblock 2021.

\bibitem[Tancik et~al.(2020)Tancik, Srinivasan, Mildenhall, Fridovich-Keil,
  Raghavan, Singhal, Ramamoorthi, Barron, and Ng]{tancik2020fourfeat}
Tancik, M., Srinivasan, P.~P., Mildenhall, B., Fridovich-Keil, S., Raghavan,
  N., Singhal, U., Ramamoorthi, R., Barron, J.~T., and Ng, R.
\newblock Fourier features let networks learn high frequency functions in low
  dimensional domains.
\newblock \emph{NeurIPS}, 2020.

\bibitem[Tatarchenko et~al.(2019)Tatarchenko, Richter, Ranftl, Li, Koltun, and
  Brox]{Tatarchenko_2019_CVPR}
Tatarchenko, M., Richter, S.~R., Ranftl, R., Li, Z., Koltun, V., and Brox, T.
\newblock What do single-view {3D} reconstruction networks learn?
\newblock In \emph{The IEEE Conference on Computer Vision and Pattern
  Recognition}, 2019.

\bibitem[Tretschk et~al.(2020)Tretschk, Tewari, Golyanik, Zollh\"{o}fer, Stoll,
  and Theobalt]{Tretschk2020PatchNets}
Tretschk, E., Tewari, A., Golyanik, V., Zollh\"{o}fer, M., Stoll, C., and
  Theobalt, C.
\newblock {PatchNets: Patch-Based Generalizable Deep Implicit 3D Shape
  Representations}.
\newblock \emph{European Conference on Computer Vision}, 2020.

\bibitem[Wang et~al.(2018)Wang, Zhang, Li, Fu, Liu, and Jiang]{WangZLFLJ18}
Wang, N., Zhang, Y., Li, Z., Fu, Y., Liu, W., and Jiang, Y.
\newblock Pixel2mesh: Generating 3{D} mesh models from single {RGB} images.
\newblock In \emph{European Conference on Computer Vision}, pp.\  55--71, 2018.

\bibitem[Wang et~al.(2019{\natexlab{a}})Wang, Ceylan, Mech, and
  Neumann]{wang20193dn}
Wang, W., Ceylan, D., Mech, R., and Neumann, U.
\newblock 3{DN}: 3{D} deformation network.
\newblock In \emph{{IEEE} International Conference on Computer Vision},
  2019{\natexlab{a}}.

\bibitem[Wang et~al.(2019{\natexlab{b}})Wang, Xu, Ceylan, Mech, and
  Neumann]{xu2019disn}
Wang, W., Xu, Q., Ceylan, D., Mech, R., and Neumann, U.
\newblock {DISN}: Deep implicit surface network for high-quality single-view
  3{D} reconstruction.
\newblock In \emph{NeurIPS}, 2019{\natexlab{b}}.

\bibitem[Wen et~al.(2020{\natexlab{a}})Wen, Han, Liu, and Liu]{9187572}
Wen, X., Han, Z., Liu, X., and Liu, Y.-S.
\newblock Point2spatialcapsule: Aggregating features and spatial relationships
  of local regions on point clouds using spatial-aware capsules.
\newblock \emph{IEEE Transactions on Image Processing}, 29:\penalty0
  8855--8869, 2020{\natexlab{a}}.

\bibitem[Wen et~al.(2020{\natexlab{b}})Wen, Li, Han, and Liu]{wenxin_2020_CVPR}
Wen, X., Li, T., Han, Z., and Liu, Y.-S.
\newblock Point cloud completion by skip-attention network with hierarchical
  folding.
\newblock In \emph{IEEE Conference on Computer Vision and Pattern Recognition},
  2020{\natexlab{b}}.

\bibitem[Wen et~al.(2021{\natexlab{a}})Wen, Han, Cao, Wan, Zheng, and
  Liu]{wenxin_2021b_CVPR}
Wen, X., Han, Z., Cao, Y.-P., Wan, P., Zheng, W., and Liu, Y.-S.
\newblock Cycle4completion: Unpaired point cloud completion using cycle
  transformation with missing region coding.
\newblock In \emph{IEEE Conference on Computer Vision and Pattern Recognition},
  2021{\natexlab{a}}.

\bibitem[Wen et~al.(2021{\natexlab{b}})Wen, Xiang, Han, Cao, Wan, Zheng, and
  Liu]{wenxin_2021a_CVPR}
Wen, X., Xiang, P., Han, Z., Cao, Y.-P., Wan, P., Zheng, W., and Liu, Y.-S.
\newblock Pmp-net: Point cloud completion by learning multi-step point moving
  paths.
\newblock In \emph{IEEE Conference on Computer Vision and Pattern Recognition},
  2021{\natexlab{b}}.

\bibitem[Williams et~al.(2019)Williams, Schneider, Silva, Zorin, Bruna, and
  Panozzo]{Williams_2019_CVPR}
Williams, F., Schneider, T., Silva, C., Zorin, D., Bruna, J., and Panozzo, D.
\newblock Deep geometric prior for surface reconstruction.
\newblock In \emph{IEEE Conference on Computer Vision and Pattern Recognition},
  2019.

\bibitem[Wu \& Sun(2020)Wu and Sun]{DBLP:journals/cgf/WuS20}
Wu, Y. and Sun, Z.
\newblock {DFR:} differentiable function rendering for learning {3D} generation
  from images.
\newblock \emph{Computer Graphics Forum}, 39\penalty0 (5):\penalty0 241--252,
  2020.

\bibitem[Zakharov et~al.(2020)Zakharov, Kehl, Bhargava, and
  Gaidon]{prior2019SDFRcvpr}
Zakharov, S., Kehl, W., Bhargava, A., and Gaidon, A.
\newblock Autolabeling 3{D} objects with differentiable rendering of sdf shape
  priors.
\newblock In \emph{IEEE Conference on Computer Vision and Pattern Recognition},
  2020.

\end{thebibliography}
\bibliographystyle{icml2021}

\end{document}